\documentclass[lettersize,journal]{IEEEtran}
\usepackage{cite}
\usepackage{booktabs}
\usepackage{booktabs}
\usepackage{hyperref} 
\usepackage{multirow}
\usepackage{colortbl}
\usepackage{xcolor}
\usepackage{tabularx}
\usepackage{dblfloatfix}
\usepackage{float}
\usepackage{xspace}
\usepackage{array}
\usepackage{amsmath} 
\usepackage{amssymb} 
\usepackage{bm}       
\usepackage{graphicx}
\usepackage{placeins} 
\usepackage{threeparttable}
\usepackage{tabu}
\usepackage{makecell}
\begin{document}

\title{\LARGE \bf Suite-IN++: A FlexiWear BodyNet
Integrating Global and Local Motion Features from Apple Suite for Robust Inertial Navigation}

\author{Lan Sun, \IEEEmembership{Student Member,~IEEE}, \thanks{This work was supported in part by the National Nature Science Foundation of China (NSFC) under Grant 62273229, and in part by ByteDance project under Grant CT20250103117914. Lan Sun, Songpengcheng Xia, Jiarui Yang, Ling Pei are with Shanghai Key Laboratory of Navigation and Location-based Services, School of Electronic Information and Electrical Engineering, Shanghai Jiao Tong University, Shanghai, China, 200240. E-mail: lan.sun; songpengchengxia; jr.yang; ling.pei@sjtu.edu.cn. (Corresponding author: Ling Pei, Songpengcheng Xia.)}
	Songpengcheng Xia\IEEEauthorrefmark{1}, \IEEEmembership{Member,~IEEE},
    Jiarui Yang, \IEEEmembership{Student Member,~IEEE},
	Ling Pei\IEEEauthorrefmark{1},  \IEEEmembership{Senior Member,~IEEE},\\

}

\markboth{IEEE TRANSACTIONS ON MOBILE COMPUTING,~Vol.~14, No.~8, August~2025}%
{Shell \MakeLowercase{\textit{et al.}}: A Sample Article Using IEEEtran.cls for IEEE Journals}


\maketitle

\begin{abstract}

The proliferation of wearable technology has established multi-device ecosystems comprising smartphones, smartwatches, and headphones as critical enablers for ubiquitous pedestrian localization. 
However, traditional pedestrian dead reckoning (PDR) struggles with diverse motion modes, while data-driven methods, despite improving accuracy, often lack robustness due to their reliance on a single-device setup. 
Therefore, a promising solution is to fully leverage existing wearable devices to form a flexiwear bodynet for robust and accurate pedestrian localization.
This paper presents Suite-IN++, a deep learning framework for flexiwear bodynet-based pedestrian localization. Suite-IN++ integrates motion data from wearable devices on different body parts, using contrastive learning to separate global and local motion features. It fuses global features based on the data reliability of each device to capture overall motion trends and employs an attention mechanism to uncover cross-device correlations in local features, extracting motion details helpful for accurate localization.
To evaluate our method, we construct a real-life flexiwear bodynet dataset, incorporating Apple Suite (iPhone, Apple Watch, and AirPods) across diverse walking modes and device configurations. Experimental results demonstrate that Suite-IN++ achieves superior localization accuracy and robustness, significantly outperforming state-of-the-art models in real-life pedestrian tracking scenarios.

\end{abstract}


\section{Introduction}
\IEEEPARstart{W}{ith} the rapid advancement of mobile computing and wearable technology, smartphones and other wearable devices have become integral to daily life \cite{9628113,9673682,sun2024suite,chen2019deep,chen2019motiontransformer}. These devices are increasingly equipped with inertial measurement units (IMUs) that enable comprehensive motion capture through various body attachments \cite{zhang2024dynamic,pei2021mars}. This technological evolution has driven significant progress in human-centric computing applications, including human pose estimation \cite{zhang2024dynamic,yi2021transpose,mollyn2023imuposer,9737726}, activity recognition \cite{xia2024timestamp,pei2021mars}, and pedestrian localization \cite{chen2019motiontransformer,herath2020ronin,wang2022inertial,herath2020ronin,10035881,jiang2024robust}.


\begin{figure}[t]
    \centering
    \includegraphics[width=0.48\textwidth]{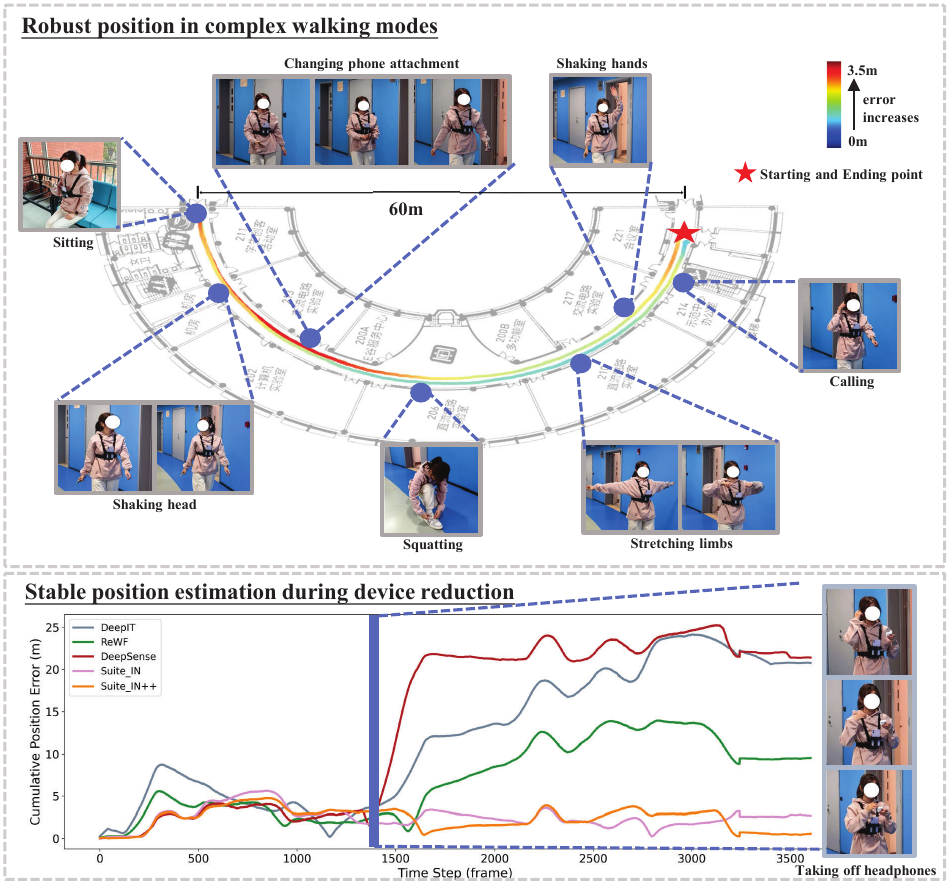} 
    \caption{Our innovative flexiwear bodynet-based approach for robust pedestrian localization: Achieving robust positioning under complex walking modes and flexible device configurations by integrating global and local motion features from a flexiwear bodynet.}
    \label{fig:1}
\end{figure}


Pedestrian dead reckoning (PDR) has attracted considerable attention for its ability to provide continuous and flexible positioning \cite{9406816,herath2020ronin,pei2018optimal,10478937}. It utilizes IMU data along with biomechanical constraints, such as zero-velocity updates \cite{10535289,10648849} and step-length priors \cite{10371336,9511443,10829797}, to estimate movement trajectories in GNSS-denied environments. Recent advances in deep learning have enabled data-driven approaches \cite{herath2020ronin, chen2018ionet} that extract high-level motion representations, significantly improving positioning accuracy\cite{chen2018ionet, chen2019deep, 8937008, herath2020ronin, yan2018ridi}. However, relying on a single sensor, such as a smartphone, for pedestrian localization makes it challenging to robustly accommodate the diverse motion modes of pedestrians in real-life scenarios.





As wearable technology becomes more widespread, utilizing multiple wearable devices to form a sensor network has emerged as a promising approach for human motion estimation \cite{9839489,9673682}, with the potential to enhance localization performance across diverse motion modes \cite{9389629,9442861,9841001,10288089}. Modern users typically carry multiple smart devices (e.g., smartphones, smartwatches, and headphones) that can collectively form a body-area sensor network \cite{jeyakumar2019sensehar,mollyn2023imuposer,10592803,10319091}. However, in real-life scenarios, natural changes in device attachment and the complexity of walking modes make the body network structure more intricate, which we define as the flexiwear bodynet (flexible wearable body network).
In a flexiwear bodynet, devices worn on different body parts (e.g., wrist, ears) capture complementary motion characteristics \cite{pei2021mars, zhang2024dynamic}: global motion features describe torso-level displacement and orientation, while local features encode limb-specific dynamics. For pedestrian localization, global features provide essential step direction and velocity estimates, whereas local features offer motion-specific details that can mitigate errors from irregular motions \cite{8868212}. Effectively integrating these complementary information sources remains a fundamental challenge in multi-device positioning systems, especially in complex real-life scenarios.

Our key insight is that wearable sensors positioned at different body parts inherently capture both global and local motion characteristics. In pedestrian localization, these two feature types play distinct roles: Global features (torso displacement, rotation, and velocity) dominate trajectory estimation, while local features (arm swing amplitude, step cadence, etc.) provide motion-specific refinements. 
Our previous work Suite-IN \cite{sun2024suite} demonstrated the potential of aggregating global motion features from multiple devices, but reveals two critical limitations:
(1) Inadequate Global Feature Aggregation: The arithmetic mean fusion of global features fails to account for device-specific information quality variations caused by body placement differences.
(2) Underutilized Local Features: The contrastive learning framework treats local motion as mere noise, disregarding its potential for enhancing position estimation.

To overcome these limitations, we extend Suite-IN \cite{sun2024suite} and introduce Suite-IN++ shown in Fig. \ref{fig:1}, which introduces two key improvements:
First, instead of averaging global features, we adopt a weighted fusion strategy that evaluates the reliability of each device’s motion information, enabling a more informed and adaptive aggregation of global motion features. Second, rather than disregarding local motion as noise, we incorporate an attention mechanism to capture intrinsic relationships between different local motion features. This allows our model to extract motion details—such as gait changes and walking modes—that enhance positioning accuracy and generalization.
In addition, unlike previous work \cite{10447298, gong2021robust,8865667}, we use real-life wearable devices (Apple Suite: iPhone, Apple Watch and Airpods) for data collection. The IMUs in these consumer-grade devices tend to have higher noise levels compared to specialized IMU systems like Xsens \cite{schepers2018xsens} and Noitom \cite{noitom2019}, introducing additional challenges for accurate pedestrian localization.

In summary, the key contributions of our paper are as follows:

\begin{itemize}

\item We propose a novel flexiwear bodynet-based pedestrian location framework, named Suite-IN++, that effectively integrates global and local motion features from wearable devices deployed on different body parts, improving the robustness and accuracy of pedestrian localization.
\item We design a contrastive learning architecture that systematically disentangles global-local motion characteristics via cross-device invariant encoding and device-specific attention modules, enabling effective coordination of complementary motion features.
\item Beyond Suite-IN, Suite-IN++ introduces two key innovations: a weighted global fusion module that adaptively adjusts device contributions according to the reliability of their motion information, while an attentive local analysis module that extracts special motion details from local features to complement global trajectory estimates.
\item We firstly present a real-life flexiwear-bodynet-based pedestrian positioning dataset \footnote{https://github.com/LannnSun/a-real-life-flexiwear-bodynet-dataset} supporting various walking modes and flexible device configurations. Compared to Suite-IN, Suite-IN++ significantly improves positioning accuracy, reducing ATE and RTE across all walking modes by up to 33.3\% and 31.86\%, respectively.



\end{itemize}

\section{Related Work}
 In this section, we review several related works on these three topics: data-driven pedestrian localization, multi-sensor fusion for wearable sensors, and contrastive learning for wearable sensors.

\subsection{Data-driven Pedestrian Localization}
Data-driven method are proposed to directly estimate position from IMU data. Data-driven smartphone inertial odometry has gained interest in recent years.
IONet \cite{chen2018ionet} is a neural network-based inertial navigation method, which uses a long short-term memory (LSTM) network model to regress pedestrian velocity magnitude and the rate of motion-heading change from smartphone data.
RIDI\cite{yan2018ridi} classifies the phone attachment by a support vector machine, and then regresses velocity for each attachment. Inertial measurements are distributed differently across domains, motiontransformer \cite{chen2019motiontransformer} exploits generative adversarial network (GAN) and domain adaptation to improve the effectiveness of inertial navigation systems for unseen domains without any paired data. 
RoNIN\cite{herath2020ronin} uses three different neural network models (LSTM, ResNet, TCN) to achieve end-to-end pedestrian positioning. Based on RIDI, it extends the application scenarios of the inertial navigation system and supports more walking modes.
In order to deploy the model on real-life devices, methods such as IMUNet\cite{zeinali2024imunet} and L-IONet\cite{8960327} are committed to developing lightweight networks to optimize network design, reduce network parameters and improve operating efficiency without affecting positioning accuracy. 

Beyond existing data-driven localization methods, recent works have focused on signal modeling and enhancement to improve feature robustness and generalization. TG-FE\cite{wang2025general} encodes inertial signals into a memory-based graph and fuses local and global information to extract transferable features. HEROS-GAN\cite{wang2025heros} enhances low-cost accelerometer signals into high-quality equivalents using optimal transport supervision and energy modulation. These approaches offer promising directions for improving inertial localization in real-world scenarios.

\subsection{Multi-sensor Fusion for Wearable Sensors}
 As wearable technology advances, there is an opportunity to leverage a variety of smart devices to enhance position estimation and broaden the applicable scenarios\cite{gong2021robust,10447298}. Multi-sensor fusion is the technique that involves gathering and combining information from multiple sensors in order to provide better information for target regression or recognition. The fusion methods can be divided into either signal, feature or decision level fusion\cite{gravina2017multi}.

By combining multi-sensor information through multi-sensor fusion, the performance of various wearable sensors-based tasks can be enhanced\cite{gong2021robust,10447298,jeyakumar2019sensehar,xue2019deepfusion,yao2017deepsense}. Gong et al.\cite{gong2021robust}proposes a multi-sensor fusion pipeline called DeepIT, which integrates IMU measurements of smartphones and associated earbuds through a reliability network to achieve inertial tracking. Restrained-Weighted-Fusion is introduced by Song et al.\cite{10447298} to enhance fusion accuracy and robustness of multi-node fusion positioning, and gumbel softmax resampling is used to optimize the weight of each sensor. Sensehar\cite{jeyakumar2019sensehar} improves human activity recognition performance by extracting features shared by multiple sensors to represent multi-sensor fusion feature. DeepSense\cite{yao2017deepsense} is a unified model to fuse multiple similar sensors to solve both classification and regression problems such as human activity recognition and motion tracking. DeepFusion\cite{xue2019deepfusion} further considers the fusion weights and cross-sensor correlations of different sensors, complementarily utilizing multi-sensor information.

\begin{table*}[ht]
\centering
\caption{Inertial Navigation Datasets.} 
\resizebox{\textwidth}{!}{  
\begin{tabular}{cccccc}  
\toprule
Dataset & Seqs & sample rate& Device & Device Flexibility & Walking Range \\
\midrule
OxIOD   & 158 &100hz& iPhone & only change phone attachment   & small, medium \\
RoNIN   & 152 &200hz& Android phone & only change phone attachment   & medium \\
DeepIT  & /   &60hz & eSense+Android phone & flexible attachment but fixed device number            & small, medium, large \\
ours    & 429 &25\textasciitilde100hz& iPhone+iwatch+airpods & both flexible attachment and device number  & small, medium, large \\
\bottomrule
\end{tabular}
}
\begin{tablenotes}
\footnotesize
\item \textit{Small} refers to an area smaller than 30x30 $m^2$, \textit{medium} refers to an area smaller than 50x50 $m^2$, and \textit{large} refers to an area larger than 100x100 $m^2$.
\end{tablenotes}
\label{tab:1}
\end{table*}

\begin{table*}[h!]
\centering
\caption{Introduction of various walking modes.}
\begin{tabularx}{\textwidth}{lX}
\toprule
\textbf{Scenario} & \textbf{Description} \\
\midrule
\textbf{STW} & \textbf{STable Walking}. Phones remain in a relatively fixed position (e.g., handheld, in a pocket, or in a bag). Only the natural rhythmic motion of walking is present, with no deliberate changes in the phone’s orientation or placement. \\[6pt]

\textbf{PVW} & \textbf{Phone-Variation Walking}. The phone is held by hand during walking, but the way of phone attachment is changed randomly, such as switching between hands, putting it in the pocket, and putting it close to the ear to answer the call, so targeted interference is introduced into the motion signal of the phone. \\[6pt]

\textbf{MVW} & \textbf{Multi-Variation Walking}. In addition to altering the phone’s holding manner, the subject worn the devices actively introduces more complex disturbances by shaking the wrist and tilting the head, affecting all three devices simultaneously. \\[6pt]

\textbf{DRW} & \textbf{Device-Removal Walking}. During walking, one of three wearable devices (such as a watch or headphones) are removed, causing sudden and irregular disruptions in the expected signals. \\[6pt]

\textbf{DLW} & \textbf{Daily-Living Walking}. The scenario expands beyond pure walking, encompassing activities that resemble everyday life: standing still, sitting down, or squatting to tie shoelaces. This setting captures a variety of natural and routine movement patterns. \\
\bottomrule
\end{tabularx}
\label{mode describe}
\end{table*}

\subsection{Contrastive Learning for Wearable Sensors}
Multi-sensor fusion has made significant progress in feature extraction and integration. To further uncover the underlying structure and distribution of the data, contrastive learning applies unsupervised methods to deeply associate sensor features.

Contrastive learning \cite{deldari2022cocoa,liu2024focal,fortes2022learning,xia2024timestamp,10705896} has been widely explored and applied in wearable-based human-centric tasks. The core principle of contrastive learning is to acquire robust representations by distinguishing between similar and dissimilar instances, often optimized using InfoNCE loss or its variants \cite{deldari2022cocoa}. For instance, COCOA \cite{deldari2022cocoa} leverages contrastive learning to extract high-quality representations from multi-sensor data by computing cross-correlations between different modalities while minimizing the similarity between unrelated instances. Recognizing the importance of modality-specific features in downstream tasks, Liu et al. \cite{liu2024focal} introduce an orthogonality constraint, enabling the simultaneous utilization of both modality-shared and modality-specific representations through contrastive loss.
For target modality data that lacks label, learning from the best\cite{fortes2022learning} ultilizes the contrastive representation misalignment loss between the source and target modality to extract the share feature of two modalities.
The contrastive learning in weakly supervised settings provides the supervision for unlabeled data, which bridges the gap between human activity classification and segmentation tasks.
Xia et al. \cite{xia2024timestamp} adopts the sample-to-prototype contrast module for further refining the rough activity recognition results (recognition task) in the sequence to the prediction of each sample’s activity (segmentation task).

\section{Dataset Description}

This section introduces a real-life dataset constructed using three consumer-grade wearable devices: a smartphone, a smartwatch, and a pair of headphones. To the best of our knowledge, this is the first dataset to establish a flexiwear bodynet comprising three devices for evaluating deep learning-based inertial odometry models under diverse and realistic conditions.
\begin{figure}[t]
    \centering
    \includegraphics[width=0.5\textwidth]{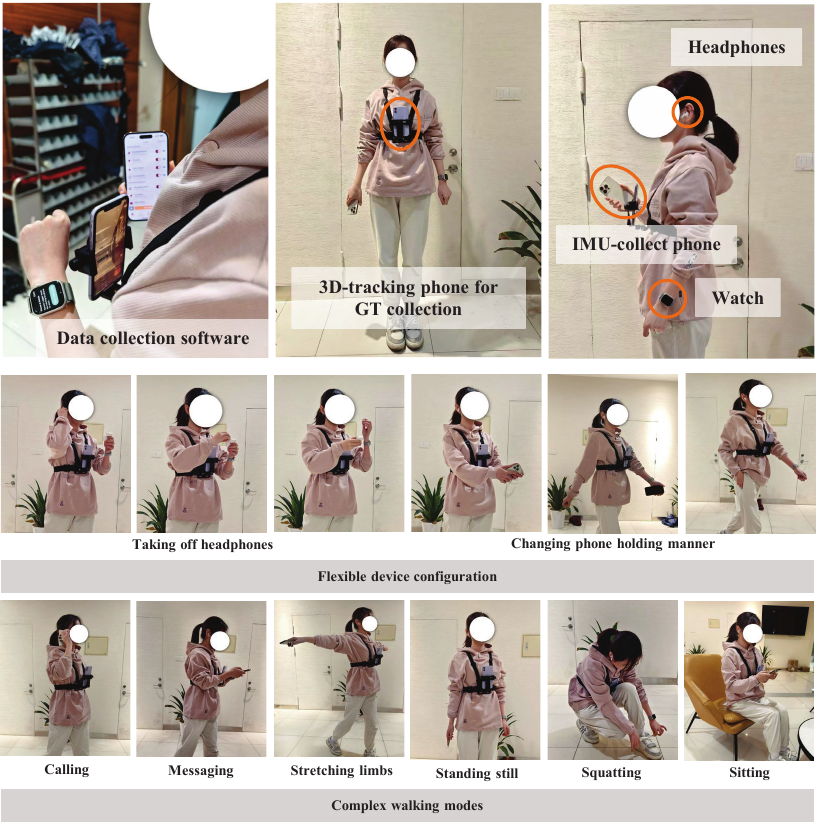} 
    \caption{Introduction to the dataset, including illustrations of device wearing, flexible device configuration, and complex walking modes.}
    \label{fig:2}
\end{figure}
\subsection{Dataset Overview}
Our dataset comprises 429 sequences, totaling approximately 20 hours of recordings and 54.5 km of walking distance, collected from 12 participants across 14 different scenes spanning two cities and four buildings.
To simulate realistic usage conditions, participants were instructed to wear devices in various configurations (e.g., handheld, in-pocket, in-backpack) and perform diverse daily walking behaviors (e.g., removing a device, sitting down, standing still, squatting mid-walk).
Compared with existing mainstream data sets such as OxIOD\cite{chen2020deep}, RoNIN\cite{herath2020ronin} and DeepIT\cite{gong2021robust}, our dataset is significantly larger, covers a broader range of walking areas and device configurations, and captures more complex motion patterns. The comprehensive dataset statistics are presented in Tab. \ref{tab:1}, where analysis of the RoNIN dataset is restricted to its publicly accessible portion due to limited data availability constraints


To capture varying levels of motion complexity, we define five walking modes, summarized in Tab. \ref{mode describe}. The simplest mode, \textbf{STW}, involves stable and consistent walking, while subsequent modes (\textbf{PVW}, \textbf{MVW}) introduce motion disturbances to one or multiple devices. The most challenging modes (\textbf{DRW}, \textbf{DLW}) simulate device removal and incorporate everyday activities. This progressive structure makes the dataset well-suited for modeling real-life pedestrian motion. The diversity in device configurations and walking modes also makes it applicable beyond localization, including tasks such as human action recognition \cite{xia2024timestamp, yao2017deepsense, xue2019deepfusion} and pose estimation \cite{zhang2024dynamic, mollyn2023imuposer, 9737726}.

\subsection{Data Collection Apparatus}
Our data collection apparatus consisted of a smartphone (Apple iPhone 14 Pro), a smartwatch (Apple Watch Series 8) on the left wrist, and one pair of headphones (Apple AirPods 3) worn in the ears. 
IMU data were collected at 100 Hz for both the iPhone and Apple Watch, and 25 Hz for the AirPods.  To ensure synchronized data acquisition, all IMU streams were downsampled to 25 Hz, consistent with the AirPods’ maximum rate.
We use the Sensor Logger \footnote{https://github.com/tszheichoi/awesome-sensor-logger} of the iOS system for data collection, with the Apple Watch and AirPods transmitting IMU data to the iPhone via Bluetooth, which then forwards it to a laptop for further processing. 

To obtain ground-truth trajectories, we used ARKit \footnote{https://developer.apple.com/documentation/arkit/}, a tightly-coupled filtering-based visual-inertial odometry (VIO) framework integrated into iOS. ARKit achieves a drift error of approximately 0.02 m per second \cite{kim2022benchmark}. 
We use a harness to attach a 3D tracking phone (iPhone 11) to a body to obtain the ground truth position frome ARKit and let subjects handle the other phone freely for IMU data collection. The 3D tracking phone records pose estimates at 30 Hz, represented by a translation vector and a unit quaternion. Ground-truth is only collected for the 3D tracking phone attached to a harness, as our objective is to estimate the human body's trajectory rather than that of the device movement. Fig. \ref{fig:2} shows how the devices are worn and the specific settings of the dataset: flexible device configuration and complex walking modes.

\subsection{Data Processing}
Before the data-collection, we performed sensor bias calibration and spatial alignment between the data-collection phone and tracking phone. 
The rotation matrix at the initial timestamp was used to project IMU readings from the phone into the global coordinate system defined by VIO. The smartwatch and headphones were natively aligned with the data-collection phone during acquisition.
Accurate temporal synchronization between VIO and IMU is crucial due to the high-frequency and time-sensitive nature of inertial data.
Following the protocol from \cite{li2021motion}, participants were instructed to jump three times at the beginning and end of each session. These sharp vertical spikes were used to align timestamps by matching peak patterns between VIO and IMU sequences.
\begin{figure*}[t]
    \centering
    \includegraphics[width=0.9\textwidth]{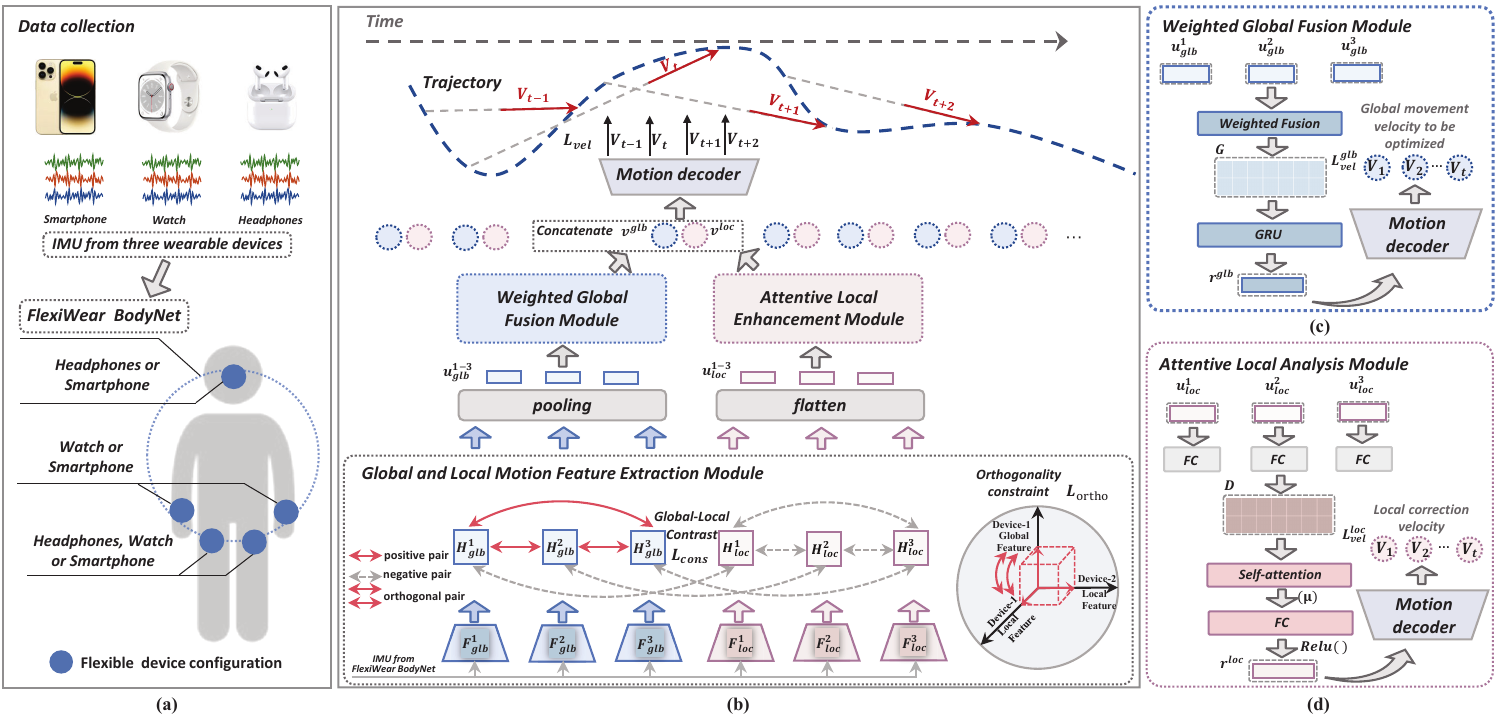} 
    \caption{Overview of our flexiwear-bodynet-based positioning framework. (a) shows the device and flexiwear bodynet used for data collection. (b) provides an overview of the Suite-IN++ algorithm, which consists of three key modules: (1) global and local motion feature extraction, (2) a weighted global fusion module (detailed in (c)), and (3) an attentive local analysis module (detailed in (d)). These modules collectively aggregate motion information from the flexiwear bodynet while distinguishing between global and local motion, enhancing the accuracy and consistency of position estimation.}
    \label{3}
\end{figure*}


\section{Method}
\subsection{Problem Statement}
\label{PS}

In this article, we define the $D$-dimensional wearable sensory sequence of length $T$ as $\mathbf{X_{1:T}} = [\mathbf{x}_1, \ldots, \mathbf{x}_T]$, where $\mathbf{x}_t \in \mathbb{R}^D$. Given $J$ wearable devices in our dataset, each sensory input at the $t$-th time step can be represented as $\mathbf{x}_t = [\mathbf{x}^1_t, \ldots, \mathbf{x}^J_t]$. Specifically, for the $j$-th wearable device, the acceleration and angular velocity at time step $t$ are denoted as $\mathbf{x}^j_t = [\mathbf{a}\ \boldsymbol{\omega}] \in \mathbb{R}^6$.
For pedestrian localization, we segment the sequence of length $T$ into overlapping windows using a sliding window approach. With a predefined window stride, the sensory sequence can be divided into $N$ windows. To simplify notation, the input data for window $n$ with a length of $L$ is represented as $\mathbf{X}_n = [\mathbf{x}t, \ldots, \mathbf{x}{t+L-1}]$, where $n \in [{1, \ldots, N}]$.

Our objective is to estimate the mean velocity $\mathbf{v}_n = [\mathbf{v}_x\ \mathbf{v}_y] \in \mathbb{R}^2$ within each window. By integrating the estimated velocities $\mathbf{v}_n$ across all windows, we can reconstruct the subject’s trajectory. In addition, we extract intermediate motion features from each sensor modality: global motion features $\mathbf{H}^{j}_\mathrm{glb}$ and local motion features $\mathbf{H}^{j}_\mathrm{loc}$. The global features primarily capture overall motion trends, playing a dominant role in trajectory estimation, while the local features encode motion-specific nuances, refining the trajectory with finer details. These two types of features complement each other, enhancing position estimation accuracy. As shown in Fig. \ref{3}, our proposed model comprises three technical modules: 1) decouple of global and local motion features for trajectory regression; 2) weighted global fusion for overall motion trends capture; 3) attentive local analysis for motion details acquisition.


\subsection{Decouple of global and local motion features for trajectory regression}

Wearable devices are typically worn on different parts of the body, and due to the irregular nature of limb movements, the motion data captured by these devices often contain complex information. Specifically, the motion information reflecting the overall motion trend of the human body is embedded in the global motion representation shared within multi devices, while each device-specific latent representation captures the local subtle motion changes. 
We believe that during human movements, the motion features captured by each device are composed of both global features and local features. The global features reflect the overall motion state of the body in space, such as displacement, rotation, and velocity, whereas the local features capture the fine-grained changes at each device’s location (such as the amplitude and rhythm of wrist swings). To handle the heterogeneity between global and local motion perception, we separate their representations effectively and leverage effective motion information to achieve velocity estimation and trajectory regression, as shown in Global and Local Motion Feature Extraction Module in Fig. \ref{3} (b).

\textbf{Independent wearable sensor feature extraction:}
As outlined in Sec.\ref{PS}, our model takes sensor data $\mathbf{X} = [\mathbf{X}^1,...,\mathbf{X}^J]$ from $J$ wearable devices as input, the data of $j$-th wearable device is denoted as $\mathbf{X}^j$. We use independent feature extractors $\mathrm{F}^j_\mathrm{glb}(\cdot)$ and $\mathrm{F}^j_\mathrm{loc}(\cdot)$ to extract global features $\mathbf{H}^j_\mathrm{glb}$ and local features $\mathbf{H}^j_\mathrm{loc}$ from each independent sensory data, represented by
\begin{equation}
\mathbf{H}^j_\mathrm{glb} = \mathrm{F}^j_\mathrm{glb}(\mathbf{X}^j)
\end{equation}
and 
\begin{equation}
\mathbf{H}^j_\mathrm{loc} = \mathrm{F}^j_\mathrm{loc}(\mathbf{X}^j).
\end{equation}
The independent feature extractors $\mathrm{F}^j_\mathrm{glb}(\cdot)$and$\mathrm{F}^j_\mathrm{loc}(\cdot)$ have the same structure composed of six 1D CNN blocks but do not share weights. In our model, the activation function ReLU, batch normalization and dropout technique are leveraged in CNN blocks, with max pooling along the temporal and sensor channel dimension to extract motion features from the wearable devices. It is worth noting that the convolution kernel size is $[3,2,2,2,2,2]$ along the temporal dimension, ensuring that the convolution slides over the time dimension to capture temporal patterns without altering the sensor channel dimensionality \cite{pei2021mars,10899893}.

\textbf{Contrastive learning for global and local motion feature separation:} In complex walking modes, global features capture the overall movement trend, while local features reflect subtle dynamics. Distinguishing between them is crucial for creating highly discriminative motion representations. Contrastive learning uses the positive and negative sample pairs to enable the model to automatically strengthen the difference between the two types of features in self-supervised training, thereby effectively distinguishing and capturing global and local information. For the same sensory data, our model iterates over all extracted features, regarding global features of different modalities  $(\mathbf{H}^i_\mathrm{glb},\mathbf{H}^j_\mathrm{glb})$ as positive pairs, regarding global feature and local feature of each modality $(\mathbf{H}^j_\mathrm{glb},\mathbf{H}^j_\mathrm{loc})$ as negative pairs, and regarding local features between different modalities $(\mathbf{H}^i_\mathrm{loc},\mathbf{H}^j_\mathrm{loc})$ as negative pairs. We calculate the constractive loss following InfoNCE loss\cite{tian2020contrastive, liu2024focal} with these positive and negative pairs, represesnted by:

\begin{equation}
\scalebox{0.85}{$ 
\mathcal{L}_\mathrm{con} = -\sum_{\substack{i,j \in [1,\ldots,J] \\ i \neq j}} 
\log \frac{s\left( \mathbf{H}^i_\mathrm{glb}, \mathbf{H}^j_\mathrm{glb} \right)}
{s\left( \mathbf{H}^i_\mathrm{glb}, \mathbf{H}^j_\mathrm{glb} \right) 
+ S\left( \mathbf{H}^i_\mathrm{glb}, \mathbf{H}^j_\mathrm{loc} \right) 
+ S\left( \mathbf{H}^i_\mathrm{loc}, \mathbf{H}^j_\mathrm{loc} \right)}
$},
\label{lcon}
\end{equation}
where
\begin{equation}
\scalebox{0.85}{$
\left\{  
             \begin{array}{lr}  
\small
             s\left (   \mathbf{H}^i_\mathrm{glb},\mathbf{H}^j_\mathrm{glb}\right )  =exp\left (   \left \langle   \mathbf{H}^i_\mathrm{glb},\mathbf{H}^j_\mathrm{glb} \right \rangle /\tau\right ) & \\  
             S\left (\mathbf{H}^j_\mathrm{glb} ,\mathbf{H}^j_\mathrm{loc}  \right )= \sum_{j=1}^{J} exp\left ( \left \langle \mathbf{H}^j_\mathrm{glb} ,\mathbf{H}^j_\mathrm{loc}  \right \rangle /\tau \right ) & \\
\displaystyle
 S\left ( \mathbf{H}^i_\mathrm{loc},\mathbf{H}^j_\mathrm{loc} \right )= \sum_{i,j\in [1,...,J],i \ne j}exp\left ( \left \langle  \mathbf{H}^i_\mathrm{loc},\mathbf{H}^j_\mathrm{loc}\right \rangle /\tau  \right ),& \\
             \end{array}  
\right. 
$}
\end{equation}
and $\left \langle \cdot  \right \rangle $ means calculating cosine similarity.

\textbf{Orthogonality constraint for global and local space:}
To prevent global motion information from contaminating the local feature space and to ensure that each wearable device captures its unique local motion patterns, we introduce orthogonality constraints. Drawing inspiration from \cite{liu2024focal, cai2024orthogonality}, we enforce these constraints both between the global and local features within the same modality and among the local features across different modalities. This design ensures that the decomposed global and local feature space captures independent semantic information, with each feature subset contributing uniquely to motion representation. To implement these constraints, we minimize the angular similarities between the corresponding feature pairs using a cosine embedding loss, which can be expressed as:
\begin{equation}
\scalebox{0.85}{$
\mathcal{L}_\mathrm{orth}=\sum_{i,j\in [1,...,J],i \ne j} \left \langle \mathbf{H}^i_\mathrm{loc},\mathbf{H}^j_\mathrm{loc} \right \rangle +\sum_{j=1}^{J}\left \langle \mathbf{H}^j_\mathrm{glb},\mathbf{H}^j_\mathrm{loc} \right \rangle.  
\label{lorth}
$}
\end{equation}

\textbf{Velocity and trajectory regression:}
After extracting global and local motion features from each wearable device, the global motion features $\mathbf{H}^j_\mathrm{glb}, j\in [1,...,J]$ can be aggregated to estimate the direction and velocity of the human body. A common aggregation method is to take the arithmetic average ($\mu$) of the global motion features of all devices\cite{jeyakumar2019sensehar,sun2024suite}. The aggregated shared features are represented as $\mathbf{\bar{H}}_\mathrm{glb}$, which fuses multi-device data into a shared low-dimensional latent space to represent the aggregated global motion features.

We take the average velocity of the window as the network's output inspired by previous works\cite{10447298, herath2020ronin}. A fully connected layer $\mathrm{FC}_\mathrm{glb}(\cdot)$ is used to simply process the aggregated global motion features to obtain the global movement velocity $\hat{\mathbf{v}}^\mathrm{glb}$:
\begin{equation}
\hat{\mathbf{v}}^\mathrm{glb} = \mathrm{FC}_\mathrm{glb}(\mathbf{\bar{H}}_\mathrm{glb}).
\end{equation}
The global movement velocity $\hat{\mathbf{v}}^\mathrm{glb}$ is supervised by the Mean-Squared-Error (MSE)\cite{koksoy2006multiresponse} loss:
\begin{equation}
\mathcal{L}_\mathrm{vel}^\mathrm{glb}=\mathrm{MSE}(\hat{\mathbf{v}}^\mathrm{glb},\mathbf{v}),
   \label{vglb}
\end{equation}
where $\mathbf{v}$ is the true value of the velocity in each window.

Our ultimate goal of this model is to obtain the human walking trajectory. Given the positon of human $\mathbf{y}_0$ at ${t}_0$, we update the velocity at each sampling moment using the window's average velocity estimated from global motion features, and then integrate over time to obtain the human walking trajectory from $t_0$, denoted as:

 \begin{equation}
\hat{\mathbf{y}}_t^\mathrm{glb}=\mathbf{y}_{t_0} +\int_{t_0}^{t} \hat{\mathbf{v}}_t^\mathrm{glb} dt.
\label{pos}
\end{equation}

The flexiwear bodynet presents two key challenges for pedestrian localization: (1) flexible device configurations and (2) complex walking modes. Relying solely on global motion features proves insufficient under these conditions, as shown in the limitations of our previous work, Suite-IN\cite{sun2024suite}: \textbf{(1) Inadequate Global Feature Aggregation:} Suite-IN employs simple arithmetic averaging to aggregate global features, disregarding the varying reliability of motion information across devices. Due to the differences in hardware quality and placement on the body, the amount of reliable global motion information carried by different devices varies. For instance, during walking, motion information captured by headphones worn on the head tends to be more stable in reflecting overall displacement. Under flexible device configurations, device attachment may change dynamically, causing the reliability of motion information from each device to fluctuate. Therefore, aggregating global features according to the reliability of each device’s motion information is crucial for capturing overall motion trends and enhancing localization robustness. \textbf{(2) Underutilized Local Features:} Suite-IN treats local features as noise, overlooking their positive contribution to motion estimation. Local features encode fine-grained motion details at each device location, such as wrist swing amplitude and rhythm, which can refine motion estimates, particularly under complex walking modes. Effectively utilizing these local features holds the potential to improve both localization accuracy and system stability.

To address these challenges, we propose two key improvements over Suite-IN: \textbf{(1)Weighted Global Fusion:} A reliability-based fusion strategy is introduced to dynamically assess the contribution of each device's motion data, enabling a more accurate aggregation of global features and ensuring robust motion trend estimation under flexible device configurations. \textbf{(2)Attentive Local Analysis:} An attention mechanism is incorporated to capture the intrinsic correlations between local motion features across devices, extracting more precise motion details to enhance localization accuracy and generalization under complex walking modes.The following sections provide a detailed explanation of these enhancements.

\subsection{Weighted global fusion for overall motion trends capture}

To address the challenge of effectively aggregating global motion features under flexible device configurations, we propose a reliability-based weighted global feature fusion strategy, as shown in Weighted Global Fusion Module in Fig. \ref{3} (c). This strategy dynamically captures changes in the reliability of motion information across devices and within each device over time due to variations in attachment. Based on these reliability assessments, the contribution of each device to the global motion features is dynamically adjusted, enabling more accurate and robust global motion feature estimation.

Inspired by\cite{xue2019deepfusion}, our model adopts a weighted fusion method to estimate the information quality (quality weight) contributed by each device and aggregate the motion features from all wearable devices in a weighted combination manner. Through this operation, our model aims to aggregate the effective global motion information contained in devices from different body parts into a shared low-dimensional latent space, so that the overall motion trend can be more robustly estimated without being affected by flexible and changeable device configurations.

 Specifically, for the global motion features $\mathbf{H}^j_\mathrm{glb}$ contained in the $j$-th wearable device, a pooling operation is first performed in the time dimension to reduce the global motion features to an appropriate scale: 

 \begin{equation}
\mathbf{u}^j_\mathrm{glb} =\mathrm{Pooling}(\mathbf{H}^j_\mathrm{glb}),
\end{equation}  
and the quality weight of the $j$-th device $e^j$ can be calculated using the following formula: 
 \begin{equation}
e^j =(\mathbf{w}^\mathrm{T}_\mathrm{glb}\mathbf{u}^j_\mathrm{glb}+{b}_\mathrm{glb})/{l}_\mathrm{glb},
\end{equation} 
where $\textbf{w}^T_{glb}$ and $b_{glb}$ are the parameters to be learned, and $l_{glb}$ denotes the length of the encoding vector $\mathbf{u}^j_{glb}$. We use a sigmoid-based function to calculate the rescaled quality weight $\tilde{\alpha}_{i}$:
 \begin{equation}
\tilde{\alpha}_{j}=\frac{\lambda_a}{1+\exp \left(-e^{j} / \lambda_b\right)}+\lambda_c,
\label{11equ}
 \end{equation}
where $\lambda_a$, $\lambda_b$ and $\lambda_c$ are the predefined hyper-parameters. The upper-bound value and lower-bound value of the rescaled weights are $\lambda_a+\lambda_c$ and $\lambda_c$, respectively. $\lambda_b$ determines the slope of the function near zero value. We can then obtain a normalized quality weight $\alpha_{j}$, as follows:
 \begin{equation}
\alpha_{j}=\frac{\tilde{\alpha}_{j}}{\sum_{j=1}^{J}\tilde{\alpha}_{j}}.
 \end{equation}
 The variance of normalized quality weights among all the devices can be reduced by setting appropriate hyperparameters. Based on the normalized quality weights of all the devices $[\alpha_{1},...\alpha_{j},...\alpha_{J}]$, our model can incorporate more devices to estimate motion, with the global combination matrix $\mathbf{G}$ computed through weighted aggregation:
  \begin{equation}
 \mathbf{G}=\sum_{j=1}^{J} \alpha_j\odot\mathbf{u}^j_\mathrm{glb}. 
  \end{equation}
To further represent sensor global combination, we applied a 2-layer stacked Gated Recurrent Unit (GRU) to finally calculate the output vector $\mathbf{r}^\mathrm{glb}$ as follows:
  \begin{equation}
\mathbf{r}^\mathrm{glb}=\mathrm{GRU}(\mathbf{G}). 
  \end{equation}
Global motion features $\mathbf{r}^\mathrm{glb}$ are processed using a fully connected layer $\mathrm{FC}_\mathrm{glb}(\cdot)$ to obtain global movement velocity $\hat{\mathbf{v}}^\mathrm{glb}$:
\begin{equation}
\hat{\mathbf{v}}^\mathrm{glb} = \mathrm{FC}_\mathrm{glb}(\mathbf{r}^\mathrm{glb}).
\end{equation}

This approach allows our model to fully leverage multi-device information by evaluating the reliability of each device’s motion data and prioritizing the more informative sources. This enables more intelligent and adaptive aggregation of global motion features, enhancing localization stability under the flexible device configurations of the flexiwear bodynet.

\subsection{Attentive local analysis for motion details acquisition}

Wearable devices are typically worn on different parts of the body, and even in the same type of motion, different wearable devices often capture different local motion information. How to make full use of local motion information to enhance global motion estimation is the key to further improve positioning accuracy, especially in complex and changeable walking modes. 
Therefore, targeting the second key aspect of the flexiwear bodynet-based localization task - flexible and changeable walking modes - we propose a attentive local analysis (Attentive Local Analysis Module in Fig. \ref{3} (d)) that aims to fully summarize the local motion information captured by wearable devices, extract richer motion details, and improve localization accuracy while maintaining a stable estimate of the overall motion trend.


Based on the global motion features and local motion features that have been obtained, the model needs to further aggregate the motion details contained in the local motion features. In this module, we first design independent linear transformations for the local motion features contained in each sensor to further extract motion information. Given the $j$-th wearable device's local motion feature $\mathbf{H}^j_\mathrm{loc}$, we adopt the flatten operation to obtain a input vector $\mathbf{u}^j_\mathrm{loc}$ for further feature extraction:
 \begin{equation}
\mathbf{u}^j_\mathrm{loc} =\mathrm{Flatten}(\mathbf{H}^j_\mathrm{loc}),
\end{equation}  
and the extracted motion information $\mathbf{d}^j$ can be expressed as:
 \begin{equation}
\mathbf{d}^j =(\textbf{w}^\mathrm{T}_\mathrm{loc}\mathbf{u}^j_\mathrm{loc}+b_\mathrm{loc})/l_\mathrm{loc}.
\end{equation} 

The local motion information $[\mathbf{d}^1,...\mathbf{d}^j...,\mathbf{d}^J]$  are then stacked as $\mathbf{D}$, a matrix containing composed local motion information. A multi-head attention mechanism is used to dynamically capture the correlation information between devices:
  \begin{equation}
\mathbf{D}^{'}=\mathrm{F}_\mathrm{attn}(\mathbf{D}), 
  \end{equation}
where $\mathrm{F}_\mathrm{attn}(\cdot)$ is the attention mechanism network, $\mathbf{D}^{'}$ is stacked by $[\mathbf{d}^{'1},...\mathbf{d}^{'j},...\mathbf{d}^{'J}]$, cross-device local motion information captured through attention mechanism. We adopt a simple approach, taking the arithmetic mean ($\mu$) of
 the cross-device local motion features $\mathbf{d}^{'}$ to aggregate local features. The aggregated local
 feature, denoted as $\mathbf{\bar{d}}^{'}$, fuses the heterogeneous sensor
 data into a shared low-dimensional latent space that better
 represents the detailed motion features.

 Finally, we further learn the interaction between sensors through non-linear transformation and obtain local motion features $\mathbf{r}^\mathrm{loc}$ containing rich motion details:
 
 \begin{equation}
     \mathbf{r}^\mathrm{loc}=\mathrm{Relu}(\mathrm{FC}(\mathbf{\bar{d}^{'}})).
 \end{equation}

\begin{table*}[t] 
	
	\centering
	\caption{\sc Localization performance of various methods under different walking modes}
	\resizebox{\textwidth}{!}{  
        		\begin{tabular}{>{\centering\arraybackslash}m{1.5cm}|>{\centering\arraybackslash}m{1cm}|>{\centering\arraybackslash}m{1.5cm}|>{\centering\arraybackslash}m{1.5cm}|>{\centering\arraybackslash}m{1.5cm}|>{\centering\arraybackslash}m{1.5cm}|>{\centering\arraybackslash}m{1.5cm}|>{\centering\arraybackslash}m{1.5cm}|>{\centering\arraybackslash}m{1.5cm}|>{\centering\arraybackslash}m{1.5cm}} 

			\toprule
			\textbf{\multirow{2}{*}{\bfseries{Test Setting}}}  & \textbf{\multirow{2}{*}{\bfseries{Metric}}} & \multicolumn{3}{c|}{\bfseries{Single-node Positioning Algorithms}} & \multicolumn{5}{c}{\bfseries{Multi-node Positioning Algorithms}}  \\
			\cmidrule{3-10} 
			& & \textbf{IONet*}\cite{chen2018ionet} & \textbf{RoNIN*}\cite{herath2020ronin} & \textbf{IMUNet*}\cite{zeinali2024imunet}  & \textbf{DeepIT*}\cite{gong2021robust}   & \textbf{ReWF}\cite{10447298} &\makecell[c]{\textbf{Deep}\\ \textbf{Sense}\cite{yao2017deepsense} }  &  \textbf{Suite-IN}\cite{sun2024suite} & \textbf{Suite-IN++}  \\ 
			\midrule
\textbf{\multirow{2}{*}{Overall}}
& ATE&4.019&3.626 &3.253&13.938&7.231&4.693&\underline{3.226}& \textbf{2.915}\\
&RTE &4.879&4.233&4.029&14.451&8.183&5.507&\underline{4.028}& \textbf{3.253}\\
			\midrule			
\textbf{\multirow{2}{*}{STW}}
&ATE &3.625&3.096&\textbf{3.040}&13.086&6.417&3.541&3.184&\underline{3.062} \\
&RTE &3.911&3.672&\textbf{3.526}&11.649&7.237&3.649&3.981& \underline{3.542}\\
			\midrule	
\textbf{\multirow{2}{*}{PVW}}
&ATE &3.102&\underline{2.303}&2.239&14.707&8.400&3.050&3.211&\textbf{2.188} \\
&RTE & 3.790&\underline{3.382}&3.459&16.235&9.574&4.106&4.235&\textbf{2.909}\\
			\midrule
\textbf{\multirow{2}{*}{MVW}}
&ATE&5.038&5.217&4.374&11.799&8.417&6.014&\underline{4.297}&\textbf{2.865}\\
&RTE &6.442&5.571&5.543&15.561&9.574&6.636&\underline{4.844}&\textbf{3.628}\\
			\midrule			
\textbf{\multirow{2}{*}{DLW}}
&ATE &4.003&4.215&3.238&17.906&8.012&5.145&\underline{3.215}&\textbf{3.147} \\
&RTE &4.955&4.686&3.873&15.708&7.946&5.022&\underline{3.464}&\textbf{3.156} \\
			\midrule	
\textbf{\multirow{2}{*}{DRW}}
&ATE &4.046&2.688&2.891&12.760&5.762&5.480&\underline{2.541}&\textbf{2.434} \\
&RTE &5.268&3.869&3.689&15.646&7.874&6.265&\underline{3.668}&\textbf{3.083}\\
			\bottomrule
	\end{tabular}}
\begin{tablenotes}
\footnotesize
\item The unit of ATE and RTE is $m$.
\end{tablenotes}	
	\label{overall}
\end{table*}
\begin{figure*}[ht]
    \centering
    \includegraphics[width=0.8\textwidth]{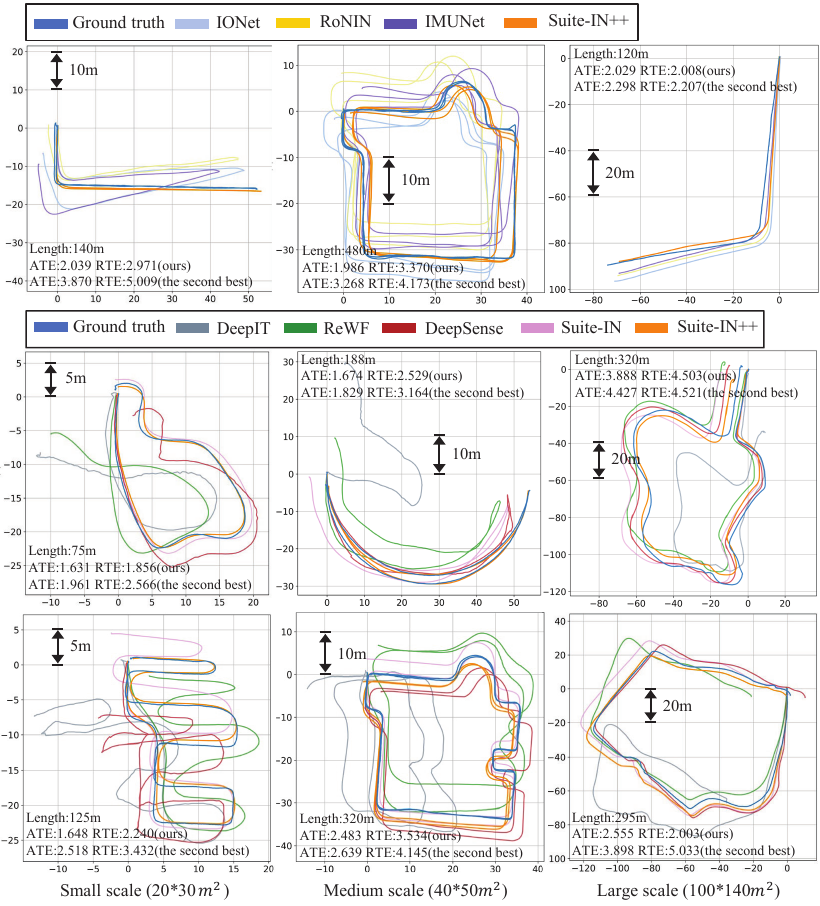} 
    \caption{Selected visualizations. We select 3 examples from each activity range, for each sequence, we label the trajectory length and report the ATE and RTE of our method and the second-best method (the unit of ATE and RTE is $m$). Our method performs best under three different ranges.}
    \label{overall_pos}
        \vspace{-5mm}
\end{figure*}

\begin{figure*}[t]
    \centering
    \includegraphics[width=\textwidth]{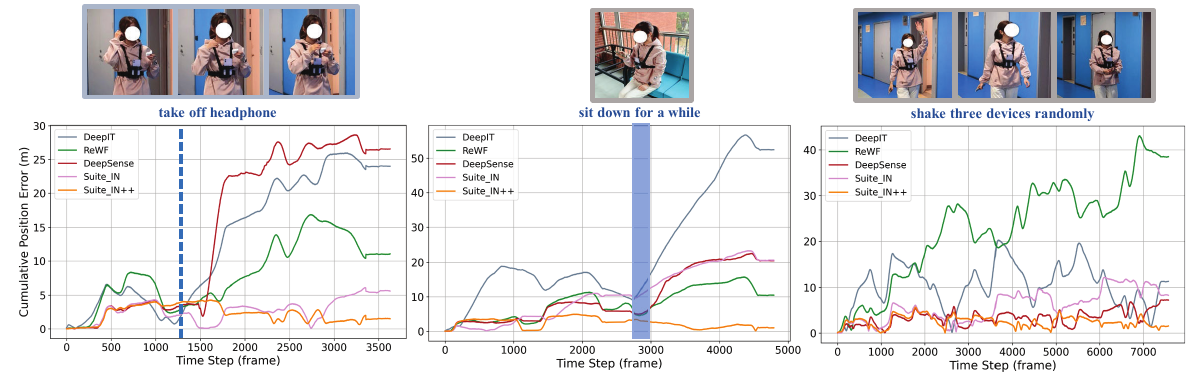} 
    \caption{Position Estimation Error. The left is the position estimation error of the sequence where the headphones are taken off midway, the middle is the sequence where the subject sit down for a while during walking, and the right is the sequence where three devices are randomly shaken.}
    \label{total_error}
    \vspace{-5mm}
\end{figure*}
By fully exploiting local motion features, our model can capture more detailed motion dynamics reflecting walking modes, user-specific characteristics, and improve localization accuracy under flexible and changeable walking modes setting in flexiwear bodynet. We use a fully connected layer $\mathrm{FC}_\mathrm{loc}(\cdot)$ to simply process local motion features to obtain local correction velocity $\hat{\mathbf{v}}^\mathrm{loc}$:
\begin{equation}
\hat{\mathbf{v}}^\mathrm{loc} = \mathrm{FC}_\mathrm{loc}(\mathbf{r}^\mathrm{loc}).
\end{equation}

Our insight is that global features dominate trajectory estimation, while local features provide motion-specific refinements. We optimize the overall velocity estimation $\hat{\mathbf{v}}$ by concatenating the local correction  velocity $\hat{\mathbf{v}}^\mathrm{loc}$ with the velocity estimated by global motion feature $\hat{\mathbf{v}}^\mathrm{glb}$ and then performing a linear transformation:
\begin{equation}
   \hat{\mathbf{v}}=\mathrm{FC}(\hat{\mathbf{v}}^\mathrm{glb} \oplus \hat{\mathbf{v}}^\mathrm{loc}).
\end{equation}
The local correction velocity $\hat{\mathbf{v}}^\mathrm{loc}$ and overall velocity $\hat{\mathbf{v}}$ is supervised by the Mean-Squared-Error (MSE) loss:
\begin{equation}
   \mathcal{L}_\mathrm{vel}^\mathrm{loc}=\mathrm{MSE}(\hat{\mathbf{v}}^\mathrm{loc},\mathbf{v}-\hat{\mathbf{v}}^\mathrm{glb})
   \label{vloc}
\end{equation}
 and
\begin{equation}
   \mathcal{L}_\mathrm{vel}=\mathrm{MSE}(\hat{\mathbf{v}},\mathbf{v}).
\label{v}
\end{equation}
Given the positon of human $\mathbf{y}_0$ at $\mathbf{t}_0$, the trajectory of human from $t_0$ can be denoted like Eq.\ref{pos}:
 \begin{equation}
\hat{\mathbf{y}}_t=\mathbf{y}_{t_0} +\int_{t_0}^{t} \hat{\mathbf{v}}_t dt.
\end{equation}

With above loss functions introduced in Eq.\ref{lcon}, Eq.\ref{lorth}, Eq.\ref{vglb}, Eq.\ref{vloc} and  Eq.\ref{v}, we set the $\lambda_\mathrm{c}$, $\lambda_\mathrm{o}$, $\lambda_\mathrm{v}^\mathrm{glb}$, $\lambda_\mathrm{v}^\mathrm{loc}$ and $\lambda_\mathrm{v}$ as hyper-parameters that determine different loss’s contribution and obtain the final loss function:
\begin{equation}
\mathcal{L}=\lambda_\mathrm{v}\cdot\mathcal{L}_\mathrm{vel}+\lambda_\mathrm{v}^\mathrm{glb}\cdot\mathcal{L}_\mathrm{vel}^\mathrm{glb}+\lambda_\mathrm{v}^\mathrm{loc}\cdot\mathcal{L}_\mathrm{vel}^\mathrm{loc}+\lambda_\mathrm{c}\cdot \mathcal{L}_\mathrm{con}+\lambda_\mathrm{o}\cdot \mathcal{L}_\mathrm{orth}.
\end{equation}

\section{Experiment}
In this section, we evaluate the proposed method and compare its performance with state-of-the-art techniques. Unless explicitly stated otherwise, all experiments share the same dataset configuration, using data exclusively from user1. The dataset is divided into training, validation, and test sets in a 6:2:2 ratio, with each set covering all walking modes included in the experiments. In the multi-sensor fusion effectiveness evaluation described in Section \ref{vc}, the test set is expanded to include two additional walking modes, PK and BG, which do not appear in the training data. This setting is designed to assess the model’s ability to generalize to unseen motion modes. In the cross-subject evaluation presented in Section \ref{ve}, the test set consists of data from four users other than user1, enabling us to evaluate the model's generalization capability across different individuals.
\subsection{Experimental Setup}
The architecture was implemented with PyTorch and trained on a NVIDIA NTX 4070 GPU. We used Adam, a first-order gradient-based optimizer\cite{kingma2014adam}, with a learning rate of 0.0001, a batch size of 128, and a window size of 100. On average, the training converged after 100 iterations. To avoid overfitting, we collected data with rich motion features and adopted Dropout \cite{srivastava2014dropout}in the network, randomly dropping 20\% of the units from the neural network during training.
We set the hyper-parameters$\lambda_v=1$, $\lambda_v^\mathrm{glb}=0.1$, $\lambda_v^\mathrm{loc}=1$, $\lambda_c=0.2$, $\lambda_o=0.05$, $\lambda_a=9$, $\lambda_b=0.01$, $\lambda_c=10$.


 We employ three standard metrics, as proposed in\cite{zhang2018tutorial}, to rigorously evaluate our results:
 
 \textbf{Absolute Trajectory Error (ATE)} signifies the cumulative error across the trajectory, represented by the Root Mean Squared Error (RMSE) between the predicted and reference trajectories. 
 
\textbf{Relative Trajectory Error (RTE)} is defined as the average RMSE between the predicted and reference trajectories over a fixed time interval.

\textbf{Cumulative Distribution Function (CDF)} is the distribution function of the probability density function of the localization error.





\subsection{Compared Algorithms}
We compare our method with the following algorithms.

\label{CM}
\textit{1)Traditional Positioning Algorithm}

\textbf{Pedestrian Dead Reckoning (PDR)\cite{tian2015enhanced}:} We utilize a step-counting algorithm to detect foot-steps and move the
position along the device heading direction by a predefined
distance of 0.67m per step.

\textit{2)Single Node Positioning Algorithms}

\textbf{IONet*\cite{chen2018ionet}:} A deep learning-based inertial navigation method employing an LSTM network model. To extend IONet for three-node positioning, we concatenate the data from three devices, referring to this variant as IONet*. Notably, the original IONet regresses distance and heading changes within each window; we adapt this to estimate velocity, ensuring consistency with our method.

\textbf{RoNIN*\cite{herath2020ronin}:} A deep learning-based inertial tracking method that employs three different backbones (LSTM, ResNet, TCN), with ResNet achieving the highest accuracy. We extend the official RoNIN (ResNet) implementation by applying a concatenation approach for three-node data fusion, denoted as RoNIN*.

\textbf{IMUNet*\cite{zeinali2024imunet}:} IMUNet introduces a one-dimensional version of the state-of-the-art convolutional neural network (CNN) network for inertial position estimation on the edge device implementation. We extend it to three-node positioning through concatenation, named as IMUNet*.     

\textit{3)Multi-Node Positioning Algorithm}

\textbf{DeepIT*\cite{gong2021robust}:}  An inertial navigation method that integrates smartphone and headphone data using an LSTM. We extend it with DeepIT* to fuse data from three sensors with primal weighting and modify the regression target from distance and heading to velocity for consistency with our method.

\textbf{ReWF\cite{10447298}:} A three-node localization method utilizing inertial sensors, comprising a ResNet-based inertial encoder and an LSTM-based sensor weight extractor. As the code is not publicly available, we implement the ReWF1 algorithm locally.

\textbf{DeepSense*\cite{yao2017deepsense}:} DeepSense is the classic learning model for HAR of multi-sensor data. The architecture of DeepSense includes three layers of local CNN, three layers of global CNN and two layers of GRU. We implement DeepSense* for positioning based on the settings in the original article.

\textbf{Suite-IN\cite{sun2024suite}:} Our previous work, a three-node positioning method based on wearable devices that uses shared global motion information contained in multiple devices.

\begin{figure}[t]
    \centering
    \includegraphics[width=0.49\textwidth]{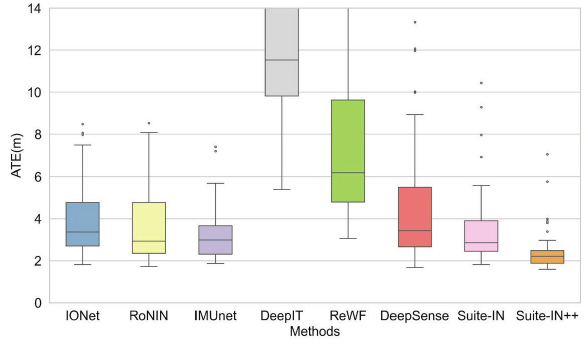} 
    \caption{Qualitative results of ATE box plot for seven competing methods and ours.}

    \label{box_ate}
  
\end{figure}

\begin{figure}[t]
    \centering
    \includegraphics[width=0.48\textwidth]{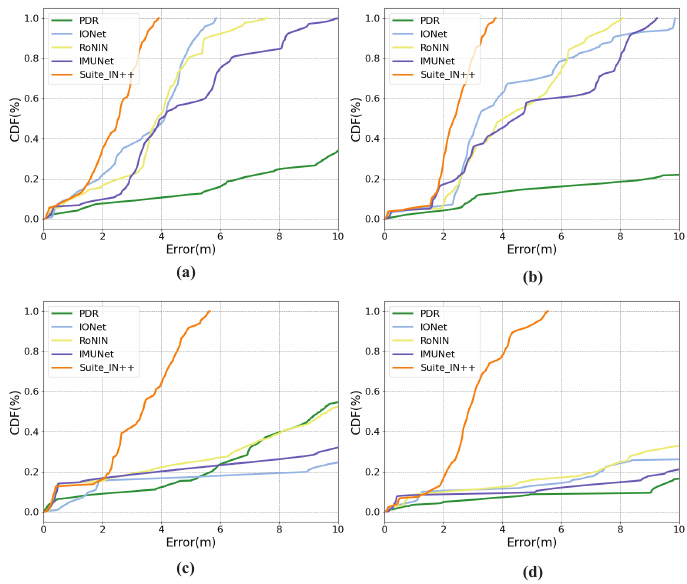} 
    \caption{ CDF of different methods under different smartphone holding manners, (a) Handheld, (b) Multi-changed Phone hold manner, (c) Phone in Pocket, (d) Phone in Bag.
}
    \label{cdf}
\end{figure}

\subsection{Comparison with the State-of-the-Art methods}
\label{vc}
For fair and meaningful evaluation, we trained all competing models on the same setting, and compared their performance to our model. 
This section presents three experiments: (1) comparing multi-node localization performance between our method and other approaches across various walking modes, (2) evaluating localization performance under flexible device configurations against multi-node localization methods, and (3) comparing our method with single-node localization algorithms across different walking modes.

\textbf{Overall Performance Comparison:}
Tab. \ref{overall} summarizes the localization performance of various competing algorithms. We use abbreviations such as \textbf{STW} and \textbf{PVW} to represent different walking modes and device configurations in our flexiwear-bodynet-based pedestrian positioning dataset. The specific walking modes are described in Tab. \ref{mode describe}. Experimental results show that our method outperforms multiple competing methods in different walking modes.

Specifically, in the stable walking mode \textbf{STW}, our method achieves comparable positioning results with the classic single-node localization algorithm IMUNet*\cite{zeinali2024imunet}, and outperforms multiple other localization methods. As our algorithm is specially designed to improve the positioning accuracy for complex walking modes, our method shows significant advantages in challenging scenarios (\textbf{PVW}, \textbf{MVW}, \textbf{DLW}, and \textbf{DRW}). 
Suite-IN\cite{sun2024suite}achieves suboptimal performance in overall experiment. Compared with Suite-IN, our method significantly reduces ATE and RTE in all walking modes, ATE and RTE are reduced by 0.12m (3.83\%) and 0.44m (11.03\%) for \textbf{STW}, 1.02m (31.86\%) and 1.33m (31.31\%) for \textbf{PVW}, 1.43 m (33.3\%) and 1.22 m (25.10\%) for \textbf{MVW}, 0.07 m (2.12\%) and 0.31 m (8.89\%) for \textbf{DLW}, 0.11 m (4.21\%) and 0.59 m (15.94\%) for \textbf{DRW}. 
Compared with DeepSense\cite{yao2017deepsense}, a classic multi-sensor fusion method for human motion analysis, our approach achieves significant improvements even in complex walking modes. For \textbf{MVW}, \textbf{DLW}, and \textbf{DRW}, ATE is reduced by 3.15m, 2.00m, and 3.05m, while RTE decreases by 3.01m, 1.87m, and 3.18m, respectively. These results highlight the effectiveness of our method in localization under complex motion conditions.

We compare the reconstructed trajectories of each method with the ground truth in different walking ranges and present the visual comparison results in Fig. \ref{overall_pos}. Experimental results show that our method outperforms competing methods in various walking ranges, further verifying its applicability and advantages in different walking modes. Fig. \ref{box_ate} presents a box plot of the ATE for competing models and our model under the \textbf{Overall} setting. It’s worth noting that our model not only achieves the lowest maximum ATE error but also has fewer outliers, indicating that our method consistently produces more robust performance across various walking modes. This further emphasizes the effectiveness of our approach in handling challenging and diverse location estimation tasks. 



\begin{table}[t] 
	
	\centering
	\caption{Multi-sensor fusion effectiveness verification.}
	\resizebox{0.49\textwidth}{!}{  
        		\begin{tabular}{c|c|c|c|c|c|c} 

			\toprule
		&	\textbf{Metric}  & \textbf{Overall} & \textbf{HD}&\textbf{MP}&\textbf{PK}&\textbf{BG} \\ 
			\midrule
\multirow{2}{*}{\textbf{PDR}} 
& \textbf{ATE} & 29.258 &17.174 &30.077&9.397&15.477\\ 
& \textbf{RTE} &32.470&14.338&36.383&7.711&15.841\\ \midrule
\multirow{2}{*}{\textbf{IONet}} 
& \textbf{ATE} & 13.542 &3.559 & 4.358 &\underline{ 6.438 }& 37.815 \\ 
& \textbf{RTE} & 11.304 & 5.240 & 4.486 & \underline{6.570} & 27.706 \\ \midrule
\multirow{2}{*}{\textbf{RoNIN}} 
& \textbf{ATE}
&\underline{3.945}&\underline{2.084}&\underline{2.770}&8.341&\underline{11.585}\\  
& \textbf{RTE} & \underline{4.621}& \underline{2.439}&\underline{3.492}&7.876 &\underline{13.693}\\ \midrule
\multirow{2}{*}{\textbf{IMUnet}} 
& \textbf{ATE} &4.926&2.284& 2.926&8.533&25.601\\ 
& \textbf{RTE} &5.361&2.668&3.598&8.828&23.801\\ \midrule
\multirow{2}{*}{\textbf{Suite-IN++}} 
& \textbf{ATE} & \textbf{2.859} & \textbf{1.963} &\textbf{2.478}& \textbf{4.875} & \textbf{2.694} \\ 
& \textbf{RTE} & \textbf{3.475} & \textbf{2.355} & \textbf{3.123} & \textbf{4.534} & \textbf{3.708} \\

			\bottomrule
	\end{tabular}}
\begin{tablenotes}
\footnotesize
\item The PDR, IONet, RoNIN and IMUNet are implemented based on data from smartphone, and our algorithm realize the multi-node-position based on three devices. The unit of ATE and RTE is $m$.
\end{tablenotes}
\label{phone}
\vspace{-5mm}
\end{table}

\textbf{Performance visualization for flexible device configuration:} Fig. \ref{total_error} illustrates the positioning performance of our method compared to others in scenarios with flexible device configurations and complex movements, common in real-world settings.
As shown in Fig. \ref{total_error}, when taking off the headphones or sitting down for a while during walking process, the position estimation error of our algorithm will not fluctuate abnormally. In contrast, competing methods show significant increases in estimation error under the same conditions. When the three devices shake randomly during walking, our algorithm always maintains a lower error than the comparison method, indicating that our algorithm can suppress noise interference and achieve accurate positioning in high-dynamic scenes.

\begin{table*}[t] 
	
	\centering
	\caption{\sc Ablation experiment table}
	\resizebox{\textwidth}{!}{  
        		\begin{tabular}{c|ccc|cc|cc|cc|cc|cc|cc|} 

			\toprule
			\textbf{\multirow{3}{*}{\bfseries{No.}}} &\multicolumn{3}{c|}{\bfseries{Modules}} & \multicolumn{12}{c|}{\bfseries{Data setting}} \\

			\cmidrule{2-16}
			& \multirow{2}{*}\textbf{Contrast.FE} & \multirow{2}{*}\textbf{Weighted.GF} & \multirow{2}{*}\textbf{Attentive.LA}  & \multicolumn{2}{c|}{\textbf{Overall}} & \multicolumn{2}{c|}{\textbf{STW}}  & \multicolumn{2}{c|}{\textbf{PVW}}  & \multicolumn{2}{c|}{\textbf{MVW} } &\multicolumn{2}{c|}{\textbf{DLW}} &\multicolumn{2}{c|}{\textbf{DRW }}\\ 
		\cmidrule{5-16}
 & &  & &ATE&RTE&ATE&RTE&ATE&RTE&ATE&RTE&ATE&RTE&ATE&RTE\\
 \midrule	

(1)& & & &3.229&3.627 &3.279&3.435 &2.397&3.548 &3.717&4.133 &3.689&3.828 &2.436&3.184\\
(2)& & \makecell[c]{$\surd$}& &3.016&3.524&3.068&3.543&2.380&3.485&3.573&4.228&3.161&3.363&2.455&\textbf{2.925}\\
(3)&\makecell[c]{$\surd$}& &\makecell[c]{$\surd$}&3.209&3.642&3.129&\textbf{3.392}&\textbf{2.095}&3.183&3.367&4.074&3.635&3.499&2.863&3.744\\
(4)& &\makecell[c]{$\surd$}&\makecell[c]{$\surd$}&3.383&3.740&3.352&3.633&2.387&3.222&4.113&4.124&3.569&3.477&2.978&4.085\\
(5)&\makecell[c]{$\surd$}&\makecell[c]{$\surd$}& &4.101&4.194&3.749&3.952&3.461&4.329&5.029&4.935&5.006&4.101&3.169&3.836\\
(6)&\makecell[c]{$\surd$}&\makecell[c]{$\surd$}&\makecell[c]{$\surd$}&\textbf{2.915}&\textbf{3.253}&\textbf{3.062}&3.542&2.188&\textbf{2.909}&\textbf{2.865}&\textbf{3.628}&\textbf{3.147}&\textbf{3.156}&\textbf{2.434}&3.083\\
			\bottomrule
	\end{tabular}}
	\label{ablation}
\end{table*}

\begin{figure*}[ht]
    \centering
    \includegraphics[width=0.9\textwidth]{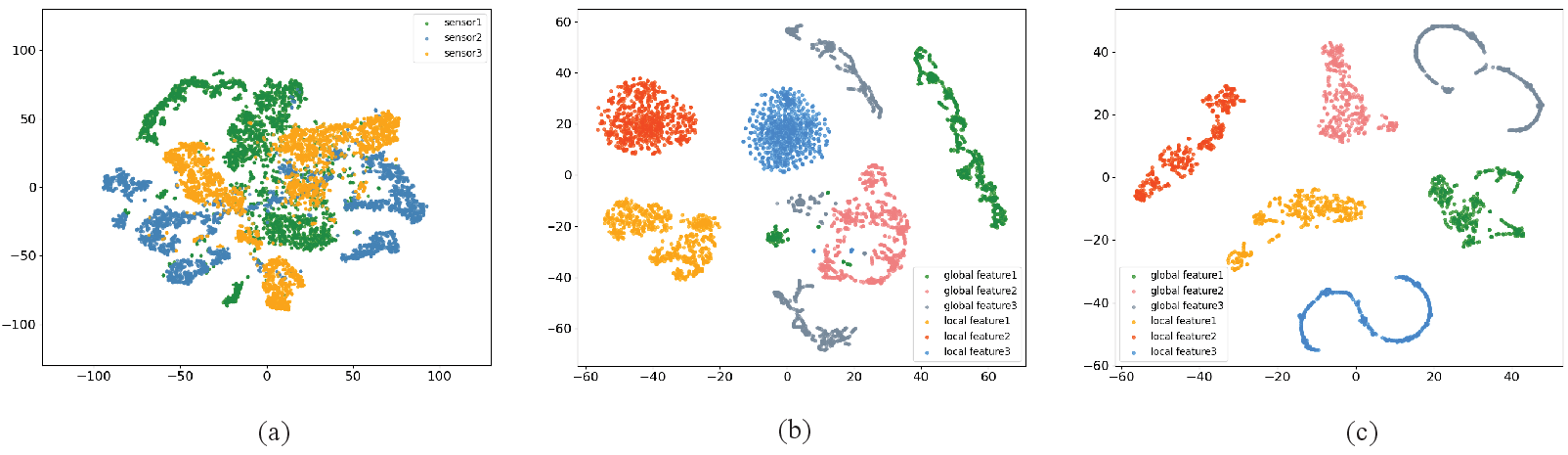} 
    \vspace{-5mm}
    \caption{ t-SNE visualization of IMU raw data and different motion features. (a) t-SNE visualization of IMU raw data from different sensors. (b) t-SNE
visualization of motion features learned in(4). (c) t-SNE
visualization of motion features learned in(6).
}
    \label{tsne}
\end{figure*}

\textbf{Multi-sensor fusion effectiveness verification:}
Tab. \ref{phone} compares the performance of our method with several state-of-the-art (SOTA) localization methods that rely only on smartphone data. In addition to the \textbf{Overall} localization accuracy, we further analyze the performance under different phone holding modes, including \textbf{HD} (\textbf{H}andhel\textbf{D}), \textbf{MP} (\textbf{M}ulti-changed \textbf{P}hone hold manner), \textbf{PK} (phone placed in \textbf{P}oc\textbf{K}et), and \textbf{BG} (phone placed in \textbf{B}a\textbf{G}). It is worth noting that \textbf{PK} and \textbf{BG} modes are not included in the training set to evaluate the generalization ability of the model.

Traditional PDR algorithms are highly dependent on step detection and step length estimation, and almost completely fail in daily localization tasks based on smartphones. Although IONet, RoNIN, and IMUNet perform well in \textbf{MP} and \textbf{HD} modes, their accuracy drops significantly in the unseen \textbf{PK} and \textbf{BG} modes. Especially in the \textbf{BG} mode, the model that relies only on the data of a single smartphone has difficulty in effectively filtering out high noise. The interference from the smartphone causes the device data quality to deteriorate, resulting in a significant increase in positioning error, exceeding 10 meters in both ATE and RTE, which is far beyond the acceptable range for daily localization applications. In contrast, our algorithm integrates data from multiple wearable devices to provide more robust positioning capabilities. Our algorithm achieves optimal performance in all phone holding manners. Even in unseen \textbf{PK} and \textbf{BG} modes, our model still maintains excellent generalization and stable positioning effects.

The cumulative error distribution function (CDF) shown in Fig. \ref{cdf} further illustrates the performance of our model. Our method outperforms the competing methods in all settings, and the maximum position error remains around 3 meters for 90\% of the test time in the complex mode \textbf{MP}. Our method improves the robustness of single-device positioning methods by integrating multi-device motion data, meeting the needs of a variety of practical application scenarios. The fusion strategy not only enhances the generalization of the model, but also achieves higher positioning accuracy in complex phone holding manners.

\subsection{Ablation study}
In this section, we examine the effectiveness of the components in our proposed method. Tab. \ref{ablation} shows the results of the ablation study on our multi-device inertial dataset in various walking modes and presents the contribution of each component (Contrast.FE: Contrastive learning based global and local motion Feature Extraction, Weighted.GF: Weighted Global Fusion, Attentive.LA: Attentive Local Analysis) in our framework. According to
the Section \ref{PS}, we compare other five kinds of variants with our proposed method: 1) We train the fundamental network by leveraging the main structure of the network to extract hybrid motion features without distinguishing between global and local motion features; 2) Based on 1), we uses the Weighted.GF module to fuse the hybrid motion features; 3) the model is trained with Contrast.FE and Attentive.LA modules based on 1); 4) based on 1), we add the
Weighted.GF and Attentive.LA modules to the model; 5) the model is trained with Contrast.FE and Weighted.GF modules based on 1) ; and 6) our approach
is trained with all modules (Contrast.FE, Weighted.GF and Attentive.LA).

The results for different components are summarized in Tab. \ref{ablation}. Our proposed method 6) consistently achieves the best localization performance across various walking modes. Method 6) outperforms variant 1), and 1) performs better than variants 3), 4), and 5), indicating that the three modules work most effectively when combined. Clearly distinguishing between global and local motion features is critical for enhancing positioning accuracy, as improper decoupling limits their complementary contribution. Moreover, the weighted fusion of hybrid motion features from multiple devices further improves performance, as demonstrated by variant 2) outperforming variant 1).

\begin{figure}[t]
    \centering  
    \includegraphics[width=0.48\textwidth]{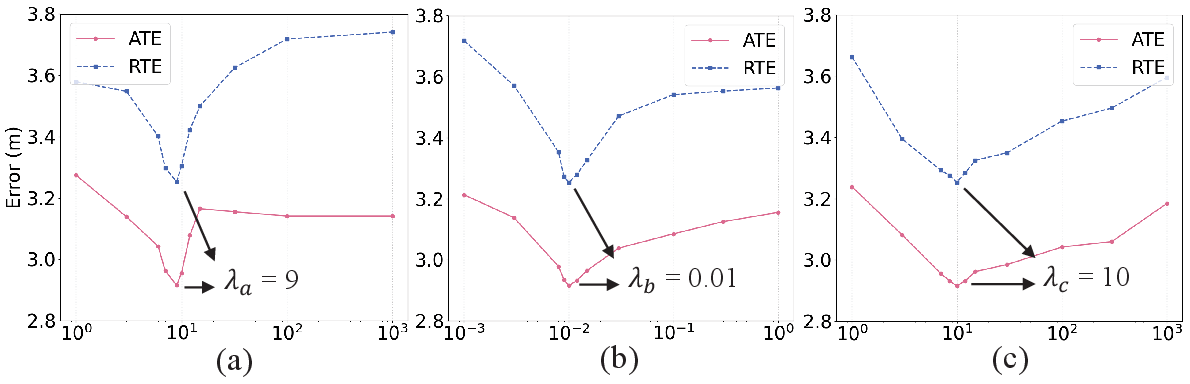} 
    \centering
    \vspace{-2mm}
    \caption{Sensitivity Analysis of ATE and RTE to Hyperparameters.(a), (b), and (c) respectively illustrate the model’s sensitivity to $\lambda_a$, $\lambda_b$ and $\lambda_c$.}
    \label{para}
\end{figure}
The results show that variant 3) surpasses variant 5), highlighting the importance of local features in positioning tasks, as they capture detailed limb motion and walking modes more accurately than global features. In complex walking modes such as \textbf{PVW}, \textbf{MVW} and \textbf{DLW}, the advantages of local features are particularly obvious, with ATE and RTE reduced by 1.366 m and 1.146 m for \textbf{PVW}, 1.662 m and 0.861 m for \textbf{MVW}, 1.371 m and 0.602 m for \textbf{DLW}, respectively, emphasizing their crucial role in enhancing positioning accuracy.

Based on variant 3), 6) introduces the Weighted.GF module to enhance global motion feature utilization. The results show that this module further improves the positioning accuracy in various walking modes, particularly in complex scenarios, as global features offer a more stable motion pattern, mitigating the negative impact of limb shaking on local features. Combining global and local motion information enables a more comprehensive capture of walking characteristics, enhancing robustness in challenging environments.
Fig.\ref{para} illustrates the sensitivity of the Weighted.GF module to the hyperparameters in Equation\ref{11equ}. When $\lambda_a$ and $\lambda_c$ are around the order of $10^1$, the model achieves an optimal balance between global trajectory consistency and local stability. When $\lambda_b$ is approximately $10^{-2}$, it effectively suppresses local pose drift and improves short-term relative positioning accuracy. Beyond these ranges, the model’s performance tends to stabilize, indicating clear parameter-sensitive regions. This parameterized design enables the model to dynamically adjust sensor weights according to different walking scenarios, achieving robust localization.

\begin{figure}[t]
    \centering
    \includegraphics[width=0.48\textwidth]{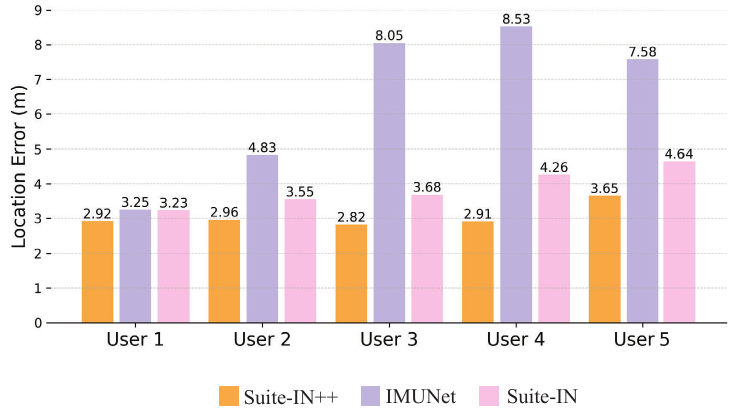} 
    \caption{Verification of the generalizability of different methods on multiple users.}
    \label{multi_user}
\end{figure}

\begin{figure}[t]
    \centering
    \includegraphics[width=0.48\textwidth]{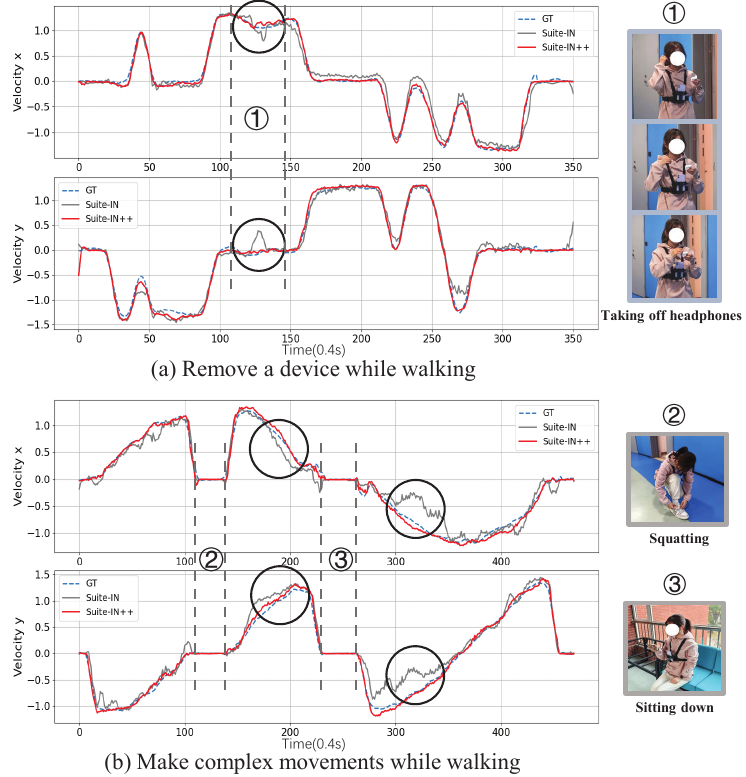} 
    \caption{Comparison of velocity estimation performance of Suite-IN++ and Suite-IN algorithms on different sequences.}
      \label{suite}
\end{figure}

Comparing variants 4) and 6), we demonstrate the effectiveness of the Contrast.FE module in the positioning task. Unlike implicit separation methods, contrastive learning enhances feature decoupling, leading to clearer separation of global and local motion features. Fig. \ref{tsne} illustrates the t-SNE projections of three different sensors' IMU raw data and the motion features in 4) and 6). The raw IMU data are relatively scattered in the latent space, and feature extraction aggregates motion features. Comparing (b) and (c) of Fig. \ref{tsne}, we can clearly see that Contrast.FE module can better aggregate motion features in the latent representation space because the features in Fig. \ref{tsne} (c) are more clustered and well-strcuted, laying a solid foundation for differentiated processing and improving overall positioning performance.

\subsection{Tests Involving Multiple Subjects}
\label{ve}
A series of experiments were conducted in different buildings with new users to show our model’s ability to generalize. Our model is trained on the data taken from user 1 and tested on different users, and all users are required to walk naturally and make natural movements such as stretching upper limbs, waving hands, shaking heads, changing the way they hold their phones, squatting to tie shoelaces, etc.

We selected the single-node positioning method IMUNet*\cite{zeinali2024imunet} and the multi-node positioning method Suite-IN\cite{sun2024suite} as benchmark comparisons to evaluate the performance advantages of our method, as they perform outstandingly in their respective categories according to Tab. \ref{overall}. As shown in Fig. \ref{multi_user}, our method can still maintain high positioning accuracy on unseen users without significant error growth, while the positioning performance of IMUNet and Suite-IN on unseen users is significantly reduced, indicating that they are less adaptable to the motion patterns of different individuals. In contrast, our method can more comprehensively model and adapt to the motion patterns of different individuals by effectively decoupling and utilizing different motion features, thereby significantly improving the generalization ability of the model.

\subsection{Comparison of Suite-IN++ and Suite-IN}
In this section, we compare two wearable-based pedestrian localization methods: Suite-IN++ and Suite-IN\cite{sun2024suite}. Unlike our previous work, Suite-IN, which treats local motion features as interference, Suite-IN++ leverages the rich motion information within these features to enhance position estimation.  

Fig. \ref{suite} presents the velocity estimation results for both methods under different walking modes. As shown in Fig. \ref{suite} (a), when the headphones are removed, the velocity estimation of Suite-IN deviates significantly from the ground truth, whereas Suite-IN++ consistently maintains accurate velocity estimation. In Fig. \ref{suite} (b), the sequence includes two stationary phases (squatting and sitting down). It is evident that after transitioning from a stationary state to walking, Suite-IN experiences a significant increase in velocity estimation error, while Suite-IN++ continues to maintain stable localization performance. This demonstrates that the motion details contained in local motion features play a crucial role in ensuring accurate localization under unstable conditions.
\begin{table}[t] 
	\centering
	\caption{Model Efficiency and Complexity of various methods.}
	\resizebox{0.49\textwidth}{!}{   
\begin{tabular}{
  >{\centering\arraybackslash}p{0.2\linewidth}|
  >{\centering\arraybackslash}p{0.2\linewidth}|
  >{\centering\arraybackslash}p{0.2\linewidth}|
  >{\centering\arraybackslash}p{0.2\linewidth}
}
\toprule
\textbf{Method} & \textbf{Inference speed (fps)} & \textbf{FLOPs (G)} & \textbf{Parameters (K)} \\ 
\midrule
\textbf{IONet}     & 44441 & 0.004 & 114.6 \\
\textbf{IONet*}    & 46036 & 0.004 & 119.2 \\
\midrule
\textbf{RoNIN}     & 9906  & 0.021 & 4438.2 \\
\textbf{RoNIN*}    & 10337 & 0.021 & 4443.7 \\
\midrule
\textbf{IMUNet}    & 9462  & 0.020 & 4068.2 \\
\textbf{IMUNet*}   & 9682  & 0.020 & 4073.6 \\
\midrule
\textbf{DeepIT*}   & 20603 & 0.005 & 317.3 \\
\textbf{ReWF}      & 5902  & 0.223 & 14049.1 \\
\textbf{DeepSense} & 17379 & 0.007 & 318.2 \\
\textbf{Suite-IN}  & 8008  & 0.026 & 104.3 \\
\textbf{Suite-IN++}& 9381  & 0.020 & 498.8 \\
\bottomrule
\end{tabular}
}
\vspace{1mm}
\begin{tablenotes}
\footnotesize
\item IONet, RoNIN, and IMUNet are originally designed based on smartphone data, whereas IONet*, RoNIN*, and IMUNet* are extended to support pedestrian localization using data from three wearable devices.
\end{tablenotes}
\label{tab:model_efficiency}
\vspace{-5mm}
\end{table}
\subsection{Model Efficiency and Deployment Analysis}
   
Although Suite-IN++ incorporates multiple CNNs, GRUs, and attention mechanisms, it was intentionally designed to support mobile deployment. To comprehensively evaluate the model's deployability, we selected three complementary indicators: (1) inference speed (fps), which reflects real-world execution efficiency; (2) FLOPs, which represent the theoretical computational complexity; and (3) the number of parameters, indicating memory and storage requirements. Together, these metrics provide a balanced and multi-dimensional assessment of the model’s suitability for mobile computing environments.

As shown in Tab. \ref{tab:model_efficiency} (Table VI in the revised manuscript), Suite-IN++ requires only 0.020 GFLOPs and 498.8K parameters, and achieves an inference speed of 9,381 fps on a desktop GPU. These results indicate low computational overhead and excellent real-time performance, making Suite-IN++ suitable for deployment on resource-constrained devices such as smartphones. This efficiency is achieved through deliberate architectural optimization: lightweight CNNs with pooling compress temporal features, GRUs and attention modules operate on low-dimensional embeddings, and the overall network maintains moderate depth to balance complexity and modeling capacity.

In contrast, RoNIN*\cite{herath2020ronin} and ReWF\cite{10447298} exhibit similar FLOPs to Suite-IN++, but their parameter counts are significantly higher—4,443.7K and 14,049.1K respectively. This results in larger model sizes and higher memory usage, which may hinder their deployment on mobile devices. Meanwhile, models like DeepSense\cite{yao2017deepsense} and DeepIT*\cite{gong2021robust} are more lightweight, but their early fusion design and shallow temporal modeling limit their ability to handle multi-sensor inputs and complex walking patterns effectively.

Moreover, compared with our previously proposed Suite-IN\cite{sun2024suite}, Suite-IN++ significantly reduces computational complexity and improves inference speed through architectural optimization, while maintaining a moderate number of parameters. By further enhancing localization accuracy, it achieves a better balance between performance and efficiency, thereby improving its practicality for real-time deployment on mobile devices.

\section{CONCLUSIONS}

This paper introduces a flexiwear bodynet-based inertial dataset covering flexible device configurations and complex walking modes, and proposes an innovative inertial positioning method based on real-life wearable devices. By extracting and integrating both global and local motion features, our method effectively captures the overall motion trend and detailed dynamics, leveraging sensor data from various parts of the body to achieve robust and high-precision pedestrian positioning.
Experimental results demonstrate that our method achieves outstanding positioning accuracy across various walking modes and device configurations. Even when users remove headphones or a smartwatch or perform natural limb movements, the model adapts effectively, maintaining stable performance despite device removal or external interference. This highlights its strong practicality for real-life applications.
Compared to our previous work Suite-IN\cite{sun2024suite}, Suite-IN++ significantly reduces ATE and RTE in various walking modes, ATE and RTE are reduced by 1.02m (31.86\%) and 1.33m (31.31\%) for \textbf{PVW}, 1.43 m (33.3\%) and 1.22 m (25.10\%) for \textbf{MVW}. 
These results highlight the importance of decoupling global and local motion information and leveraging their complementary contributions to improve positioning accuracy.
Furthermore, our innovative approach to combining global and local motion features provides a new paradigm for motion analysis in multi-device fusion, offering significant potential for future advancements in wearable-based localization systems.

\bibliographystyle{IEEEtran}
\bibliography{reference}

\newpage

\begin{IEEEbiography}[{\includegraphics[width=1in,height=1.25in,clip,keepaspectratio]{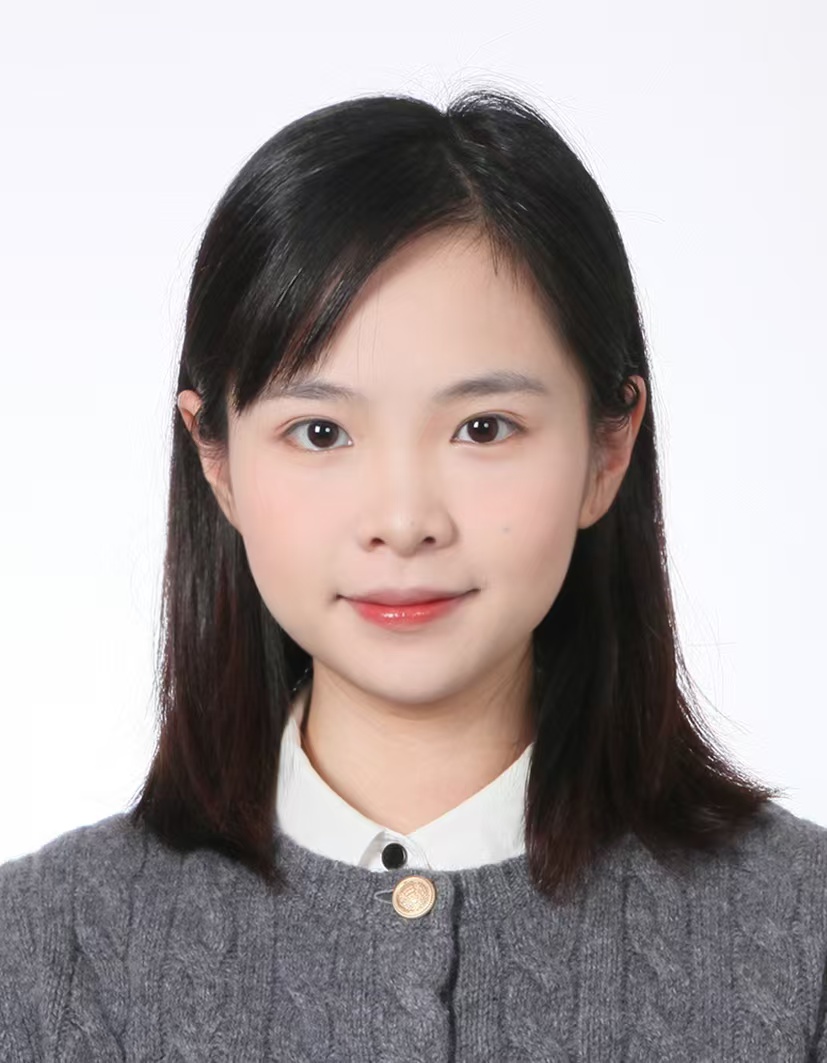}}]{Lan Sun}
(Student Member. IEEE) received the B.S. degree in Measurement and Control Technology and Instrument from Southeast University, Nanjing, China, in 2023. She is currently pursuing the Ph.D. degree with Shanghai Jiao Tong University, Shanghai, China. Her current research interests include machine learning, inertial navigation, multi-sensor fusion and wearable sensor-based human motion analysis.
\end{IEEEbiography}

\begin{IEEEbiography}[{\includegraphics[width=1in,height=1.25in,clip,keepaspectratio]{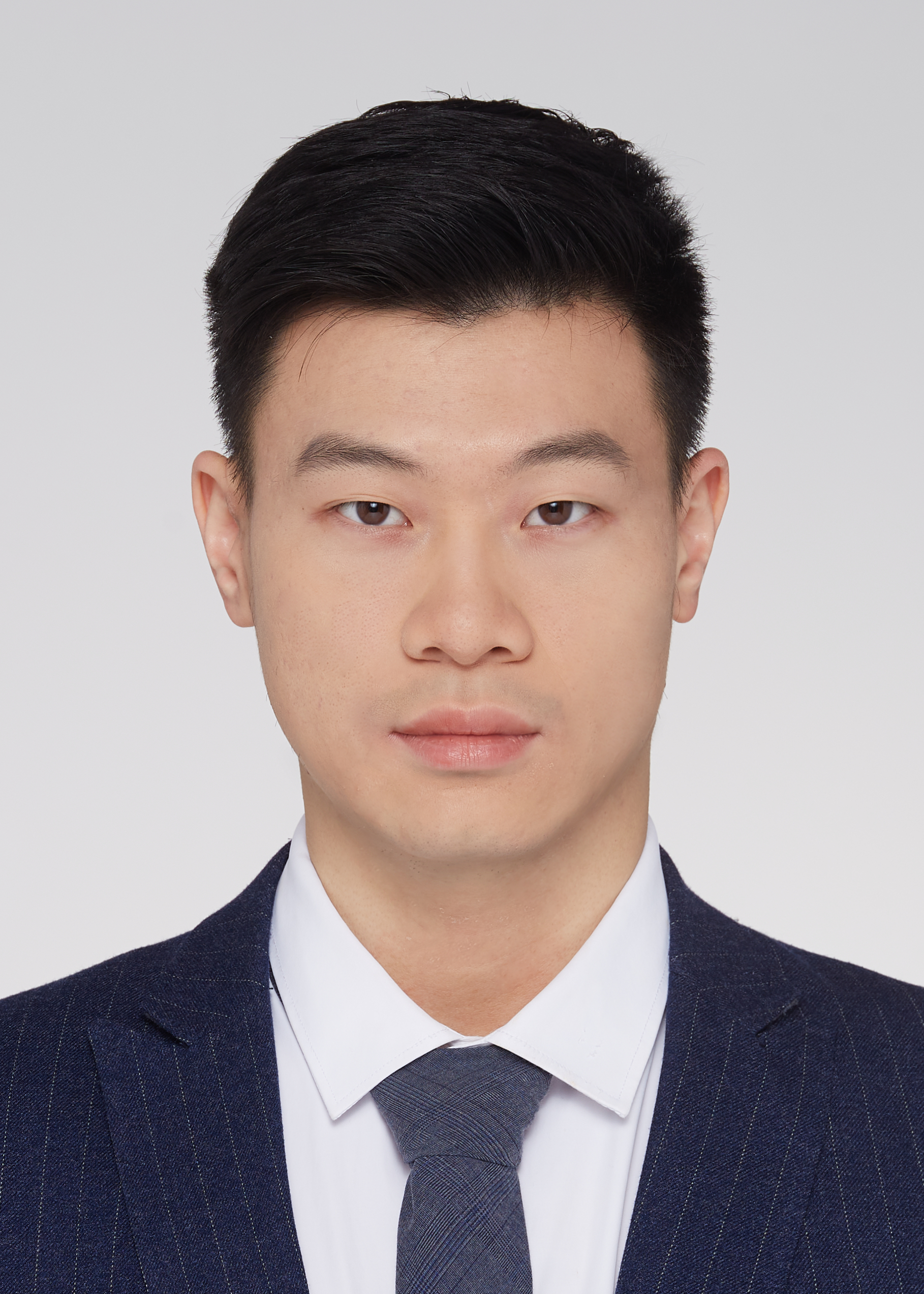}}]{Songpengcheng Xia}
		(Member. IEEE) received the B.S. degree in navigation engineering from Wuhan University, Wuhan, China, in 2019. He is currently pursuing the Ph.D. degree with Shanghai Jiao Tong University, Shanghai, China. His current research interests include machine learning, inertial navigation, multi-sensor fusion and sensor-based human activity recognition.
\end{IEEEbiography}
        \begin{IEEEbiography}[{\includegraphics[width=1in,height=1.25in,clip,keepaspectratio]{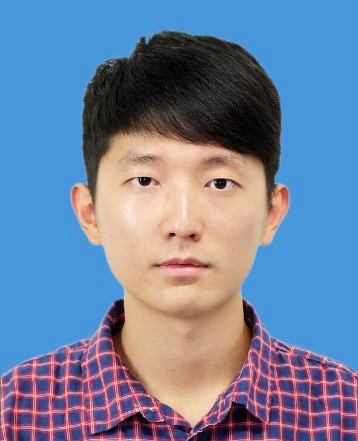}}]{Jiarui Yang}
        (Student Member, IEEE) received the B.S. degree in telecommunication engineering from Politecnico di Torino, Turin, Italy in 2018. He received the M.S. degree in communication systems from KTH Royal Institute of Technology, Stockholm, Sweden in 2021. He is currently working toward the Ph.D. degree in information and communication engineering with Shanghai Jiao Tong University, Shanghai, China. His research interests include machine learning, deep learning, and human pose estimation.

	\end{IEEEbiography}
	\begin{IEEEbiography}[{\includegraphics[width=1in,height=1.25in,clip,keepaspectratio]{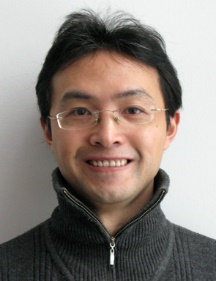}}]{Ling Pei}
		(Senior Member, IEEE) received the Ph.D. degree from Southeast University, Nanjing, China, in 2007. From 2007 to 2013, he was a Specialist Research Scientist with the Finnish Geospatial Research Institute. He is currently a Professor at the School of Electronic Information and Electrical Engineering, Shanghai Jiao Tong University. He has authored or co-authored over 100 scientific papers. He is also an inventor of 25 patents and pending patents. His main research is in the areas of indoor/outdoor seamless positioning, ubiquitous computing, wireless positioning, Bio-inspired navigation, context-aware applications, location-based services, and navigation of unmanned systems. Dr. Pei was a recipient of the Shanghai Pujiang Talent in 2014 and ranked as the World's Top 2\% scientists by Stanford University in 2022.
	\end{IEEEbiography}

\vfill

\end{document}